\documentclass[journal]{IEEEtran}
\ifCLASSINFOpdf
  \usepackage[pdftex]{graphicx}

\else
\fi
\usepackage[caption=false,font=footnotesize,subrefformat=parens]{subfig}
\usepackage{algorithmic}
\usepackage[linesnumbered,ruled,vlined]{algorithm2e}
\usepackage{amsmath}
\usepackage{amssymb}
\usepackage{amsthm}
\usepackage{graphicx}
\usepackage{booktabs}
\usepackage{cite}
\usepackage{bm}
\usepackage{float}
\usepackage{multirow}
\usepackage{color}
\usepackage{caption}
\usepackage{epstopdf}
\usepackage{hyperref}
\usepackage{fixltx2e}
\usepackage{longtable}
\usepackage{diagbox}
\usepackage{changepage}
\usepackage{comment}
\usepackage{cases}
\usepackage{makecell}
\usepackage{mathabx}
\theoremstyle{definition}

\newtheorem{definition}{Definition}

\usepackage{graphicx}
\usepackage{pifont}
\newcommand{\cmark}{\ding{51}}
\newcommand{\xmark}{\ding{55}}

\usepackage{array,threeparttable}
\theoremstyle{remark}

\theoremstyle{plain}
\newtheorem{theorem}{Theorem}
\newtheorem{lemma}{Lemma}

\begin{document}
%
\title{Multi-attribute Auction-based Resource Allocation for Twins Migration in Vehicular Metaverses: A GPT-based DRL Approach}

\author{Yongju Tong, Junlong Chen, Minrui Xu, Jiawen Kang, Zehui Xiong, Dusit Niyato, \IEEEmembership{Fellow,~IEEE}, \\Chau Yuen, \IEEEmembership{Fellow,~IEEE}, Zhu Han, \IEEEmembership{Fellow,~IEEE}}

\maketitle

\begin{abstract}
Vehicular Metaverses are developed to enhance the modern automotive industry with an immersive and safe experience among connected vehicles and roadside infrastructures, e.g., RoadSide Units (RSUs). For seamless synchronization with virtual spaces, Vehicle Twins (VTs) are constructed as digital representations of physical entities. However, resource-intensive VTs updating and high mobility of vehicles require intensive computation, communication, and storage resources, especially for their migration among  RSUs with limited coverages. To address these issues, we propose an attribute-aware auction-based mechanism to optimize resource allocation during VTs migration by considering both price and non-monetary attributes, e.g., location and reputation. In this mechanism, we propose a two-stage matching for vehicular users and Metaverse service providers in multi-attribute resource markets. First, the resource attributes matching algorithm obtains the resource attributes perfect matching, namely,  buyers and sellers can participate in a double Dutch auction (DDA). Then, we train a DDA auctioneer using a generative pre-trained transformer (GPT)-based deep reinforcement learning (DRL) algorithm to adjust the auction clocks efficiently during the auction process. We compare the performance of social welfare and auction information exchange costs with state-of-the-art baselines under different settings. Simulation results show that our proposed GPT-based DRL auction schemes have better performance than others.
\end{abstract}

\begin{IEEEkeywords}
Vehicular Metaverses, VTs migration, Multi-attribute Auction, Machine Learning, and Resource Allocation.
\end{IEEEkeywords}

%
\IEEEpeerreviewmaketitle

\section{Introduction}
\IEEEPARstart{M}{etaverses} are characterized by advanced immersive technologies such as augmented reality (AR), virtual reality (VR), mixed reality (MR), and digital twins~\cite{xu2022full,xu2023sparks}. In intelligent transportation systems, vehicular Metaverses, as the digital transformation in the automotive industry, combine immersive communication technologies in vehicular networks and leverage real-time vehicle data to offer seamless virtual-physical interaction between vehicles and their virtual representations\cite{jiang2022reliable}. Unlike high definition (HD) maps which provide accurate road information for autonomous driving\cite{he2023vi}, Vehicle Twins (VTs) serve as the digital counterparts of the physical vehicles in virtual worlds, offering extensive and precise digital reproductions spanning the entire lifecycle of vehicles, while also managing vehicular applications \cite{qian2023ofdm}. Intra-twin communication \cite{wang2023survey,bhatti2021towards}, which entails interactions between VTs and vehicles, enables vehicles to access the vehicular Metaverses via VTs to obtain vehicular Metaverses services, e.g., pedestrian detection, AR navigation, and 3D entertainment. To ensure real-time synchronization for vehicular services in vehicular Metaverses, VTs need to be continuously updated in virtual spaces, which not only enhances the seamless experience of immersion of vehicular users, including passengers and drivers, but also poses significant challenges in terms of synchronization demands\cite{verma2023survey}.

VTs in Metaverses often coexist in multiple instances. These VTs are typically resource-intensive, requiring large amounts of computation, communication, and storage resources to process VT tasks in the vehicular Metaverses~\cite{kang2024uav}. Consequently, vehicles offload VT tasks to nearby edge computing servers, located in RoadSide Units (RSUs) equipped with sufficient resources, such as bandwidth and GPUs, enabling the deployment of multiple VTs simultaneously for online interactions in virtual spaces~\cite{10185562}. RSUs process VT tasks by receiving requests from vehicles within their communication coverage, executing those VT tasks using their adequate resources, and then transmitting the results of accomplished VT tasks back to the vehicles. However, due to the limited coverage of RSUs and high mobility of vehicles, a single RSU cannot continuously provide synchronization services to vehicular users\cite{hou2023efficient}. Consequently, to maintain a consistent and seamless experience for vehicular users, each VT need to be migrated from the current RSU to the next RSU when the vehicle is going to exit the coverage of the current RSU to keep the synchronization, thus aligning the virtual space with the physical movement of vehicles\cite{9793620}.

By providing virtual assistants in virtual spaces, VTs offer immersive and personalized VT tasks like AR navigation, 3D entertainment, and tourist guidance for vehicular users~\cite{10325366}. Given the diversity of VT tasks, it is essential to consider distinct resource requirements when migrating VTs to ensure a continuously immersive experience for vehicular users. Therefore, how to achieve optimal resource allocation is challenging in vehicular Metaverses. On the one hand, RSUs acting as Metaverse service providers (MSPs) have to consume their computation, storage, and communication resources during processing VTs tasks for vehicular users\cite{hou2023uav}. On the other hand, vehicular users tend to pay different amounts of payment for multi-attribute resources according to their VT tasks. For instance, simpler tasks such as basic navigation incur lower VTs migration costs, whereas more complex operations involving advanced AR navigation or real-time data processing for entertainment or tourist guiding services result in higher fees. This variation in cost is reflective of the differing latency and resource demands related to the various types of VT tasks. Given these complexities in resource demand and cost variability for different VT tasks, it is crucial to design a mechanism that addresses the dynamic resource needs of VTs and balances the variation in cost of vehicular users and MSPs.

In this paper, we formulate the multi-attribute resource market as a bilateral market, where multiple vehicular users act as buyers to purchase required resources and multiple MSPs act as sellers to supply resources. However, it is challenging to design a trading mechanism that balances the interests of both vehicular users and MSPs, incentivizing them to participate in the multi-attribute resource market for VTs migration. To this regard, auction theory is effective for real-time and low-complexity resource allocation in non-cooperative scenarios, optimizing the social welfare through bidding and competition \cite{borjigin2018broker}, thereby balancing the supply and demand of MSPs and vehicular users. Due to the different interests between vehicular users and MSPs, we propose a double auction-based mechanism that addresses the concerns of both buyers and sellers, ensuring that the price charged to vehicular users and the payment for MSPs achieve a tradeoff.

Existing auctions that focus on mainly price are insufficient due to challenges in network connectivity and RSUs' task performance~\cite{fan2023decentralized}. Poor network quality and delays in large-scale data transfer affect the timeliness of resource allocation, while dishonest or malicious RSUs compromise reliability\cite{yu2023spatiotemporal}.
Therefore, we propose a multi-attribute double Dutch auction-based (MADDA) mechanism, which considers non-monetary attributes such as location and reputation to enhance the quality of vehicular users' experience during VT migrations.
However, considering reputation and location simultaneously complicates resource allocation, as different users have varying requirements. The auction process must evaluate multiple attributes to meet these diverse requirements, turning resource allocation and pricing into an NP-hard problem.

Fortunately, deep reinforcement learning (DRL) has emerged as a robust solution for cloud-edge resource scheduling, outperforming traditional algorithms with its adaptability to diverse and dynamic environments \cite{9759989}. However, DRL faces challenges in complex scenarios due to its reliance on limited current information \cite{9435782}, which limits its ability to deal with sequential problems. Rencently,  generative pre-trained transformer (GPT) models have demonstrated powerful sequence modeling capabilities in natural language processing, as exemplified by the outstanding performance of GPT-3 in text generation~\cite{pandey2024generative}. Several studies have integrated GPT with DRL, showing remarkable results in solving sequential modeling problems. For example, Wang \textit{et al.} proposed a novel framework combining GPT and offline RL, which significantly improved the efficiency of solving multi-step decision problems~\cite{wang2023logistics}. Inspired by these advancements, we propose a new GPT-based DRL algorithm to solve sequence problem in MADDA mechanism, specifically addressing the DDA clock sequence issue to improve auction efficiency.

Our main contributions can be summarized as follows:

\begin{itemize}
\item We design a novel generative learning-based incentive mechanism in which the learning can adapt to the environment change without prior information. The proposed mechanism combines pricing and non-monetary factors for solving the resource allocation and pricing problem of vehicle twins migration in vehicular Metaverses.
\item In this mechanism, we propose the MADDA to allocate and price VTs migration tasks between vehicular users and MSPs to achieve the seamless experience of immersion of vehicular users. The MADDA consists of: (i) resource-attributes matching by constructing Weighted Bipartite Graphs of buyers and sellers based on resource-attributes with the Kuhn-Munkres Algorithm \cite{6084856}; (ii) winner determination and pricing by DRL strategy.
\item Unlike traditional DRL, we propose a new GPT-based DRL with sequence modeling, which is more suitable for solving sequence problems (e.g., MADDA) to improve the communication efficiency of MADDA. We use the GPT-based DRL for MADDA to accelerate the allocation and pricing process of auctions, achieving near-optimal social welfare with lower information exchange cost.
\end{itemize}

The rest of this paper is organized as follows. In Section \ref{related}, we provide a review of related work. In Section \ref{system}, we describe the system. Section \ref{multi} provides the implemented details of the proposed MADDA mechanism. In Section \ref{V}, we propose a GPT-based DRL algorithm for MADDA. We illustrate the simulation experiments in Section \ref{VI} and finally conclude in Section \ref{VII}.



\begin{table*}[]
\centering
\caption{Key notation used in this paper.}
\begin{tabular}{|c|m{6cm}|c|m{6cm}|}
\hline
         \textbf{Symbol} & \centering\textbf{Definitions} & \textbf{Symbol} & \centering\textbf{Definitions}\tabularnewline
         \hline
         $\mathcal{N}$ & Set of $N$ available RSUs, regarding as MSPs & $v_{n}^{s}$ & The value function of MSPs\\ \hline
         $\mathcal{M}$ & Set of $M$ vehicular users (VUs) & $E_n$ & The spectrum efficiency of MSP $n$\\ \hline
         $P_{n}$ & Fixed position of RSUs & $f_n$ & The CPU frequency of MSP $n$'s computing chipset\\ \hline
         $O_{n}^{cp}$, $O_{n}^{com}$, $O_{n}^{s}$ & Available computation, communication, storage resources owned by RSU $n$ & $\delta_n$ & The effective capacitance coefficient of MSP $n$'s computing chipset\\ \hline
         $R_{m}^{cp}$, $R_{m}^{com}$, $R_{m}^{s}$ & The required computation, communication, and storage resources for VU $m$ to perform VT migration tasks & $\Pi_m^t$ & The transmission rate provided to VU $m$ at the time slot $t$\\ \hline
         $\mathbf{A}$ & Set of $L$ non-monetary attributes & $B_{m}$ &The bandwidth requested by VU $m$\\ \hline
         $W_{m}$ & Set of weight vector of L non-monetary attributes & $v_{m}^{b}$ & The value function of VU $m$\\ \hline
         $Q_m$ & The multi-attribute value requirement for the L attributes by VU $m$ & $\bar{T}$ & The expected latency of the VT migration task for VU $m$\\ \hline
         $q_{mi}$ & The expected attribute value of non-monetary attribute $a_{i}$ for VU $m$ & $\hat{T}$ & The maximum tolerable latency required by VUs to maintain immersion during VTs migration\\ \hline
         $\hat{Q}_n$ & The $L$ attribute values held by RSU $n$ & $\hat{\alpha}$ & Latency sensitivity factor\\ \hline
         $\hat{q}_{nj}$ & The attribute value of non-monetary attribute ${a_j}$ possessed by RSU $n$ & $g_{m}^{t}$ & The buy-bid of VU $m$ at round t\\ \hline
         $P_{m\rightarrow e}^{t-1}$ & The position of the VU $m$ establishes a communication connection with RSU $e$ at time $t-1$ & $k_{n}^{t}$ & The sell-bid of MSP $n$ at round t \\ \hline
         $d_{e\to n}^t$ & The distance between the current RSU $e$ and the next RSU $n$ at time t & $G$, $\hat{G}$ & Unweighted bipartite graph and weighted bipartite graph, respectively\\ \hline
         $D_{n}$ & The reputation value of MSP $n$ & $\Gamma $ & The resource-attributes perfect matching\\ \hline
         $s_{mn}^{k}$ & The quantity of the $k$-th resource actually provided by MSP $n$ to VU $m$ & $\hat{G}^{'}$, $\hat{G}_{l}^{'}$ & Balanced bipartite graph and equality subgraph, respectively\\ \hline
         $R_{m}^{k}$ & The quantity of the $k$-th type of resource required by VU $m$ according to $R_{m}^{cp}$, $R_{m}^{com}$ or, $R_{m}^{s}$ & $C_{B}$, $C_{S}$ & Two Dutch clocks which $C_{B}$ is buyer clock and $C_{S}$ is seller clock \\ \hline
         $E_{mn}^{k}$ & The resource feedback evaluation of VU $m$ for the $k$-th types of resource offered by MSP $n$ after transactions & $M_{B}^{t}$, $N_{S}^{t}$ & The buy-winner sequence and the sell-winner sequence at time $t$, respectively \\ \hline
         $\omega_{m}^k(a)$ & The resource weight of VU $m$ for the $k$-th resource in the $a$-th transaction & $\phi_{n}^{t}$ & The difference between the anticipated sell-bid and the actual sell-bid of seller $n$ at time $t$\\ \hline
         ${E}_{n}(a)$ & The weighted feedback evaluation of all types of resources received by MSP $n$ in the $a$-th transaction & $\phi_{m}^{t}$ & The difference between the expected buy-bid and the actual buy-bid of buyer $m$ at time $t$\\ \hline
         $t_{n}(a)$ & The transaction moment of MSP $n$ in the $a$-th transaction & $\Lambda$ & The set of candidate winning pairs which comply with the condition, i.e., $\Lambda \subseteq \Gamma$ \\ \hline
         $t_{\mu }$ & The moment when the auctioneer updates the reputation of MSP $n$ & $\hat{\Lambda}$ & The ultimate winning pairs of MADDA\\ \hline
         $\lambda_{n}(a)$ & The freshness of ${E}_{n}(a)$ & $p^{max}$ & The initialized buyer clock with the highest price\\ \hline
         $\xi $ & The parameter determining  the rate of time decay & $p^{min}$ & The initialized seller clock with the lowest price\\ \hline
         $\Psi^{t}$ & The active side of the auction clock & $p^{\ast}$ & The market clearing price\\ \hline
         $u_{m}$, $\hat{u}_{n}$ & The utility of buyer $m$ and the utility of seller $n$, respectively & $U_{B}$, $U_{S}$ & The aggregate of the ultimate buyer winner utility and the aggregate of the ultimate seller winner utility, respectively\\ \hline
         $\hat{B}_n$ & The bandwidth of RSU $n$ & $\varepsilon$ & The cost per TB of storage capacity\\ \hline
\end{tabular}
\label{table I}
\end{table*} 

\section{Related Works}\label{related}
\subsection{Vehicular Metaverses}
The concept of Metaverses, originally coined by Neal Stephenson in his science fiction novel titled \textit{Snow Crash} in 1992 \cite{wang2022metasocieties}, represents a parallel virtual world intertwined with reality \cite{9915136}. This concept has since diversified into various sectors. For example, unlike HD maps which provide accurate road information for autonomous driving\cite{he2023vi}, paper \cite{9999153} introduces the concept of vehicular Metaverses, which are defined as a future continuum between the vehicular industry and the Metaverses. Furthermore, paper \cite{9952976} discusses some opportunities and challenges of Vehicular Metaverses. Specifically, there is a focus on researching the convergence issues of edge intelligence and vehicular Metaverses. In \cite{10177684}, the author proposes a generative AI-empowered framework for the vehicular Metaverse, combining edge intelligence with an enhanced auction to synchronize autonomous vehicles and AR recommenders. 
However, this study overlooks the mobility of vehicles in vehicular Metaverses and does not address the problem of resource allocation for VT migrations.

Unlike previous works that overlook the high mobility of vehicles and the resource allocation challenges in vehicular metaverses, we propose an attribute-aware auction-based mechanism that considers the position of vehicles and RSUs to ensure efficient resource allocation during VT migrations.
\subsection{Resource Allocation for Vehicular Twins}
Optimizing resource allocation in Metaverses has become a research interest that draws significant attention for ensuring an immersive user experience in interacting with 3D objects in the virtual space~\cite{xu2023epvisa}. In \cite{10250875}, the authors present a multidimensional optimization framework for AR applications in vehicular Metaverses, aiming to balance data utility and energy efficiency in resource and transmission allocation. However, this work does not account for the mobility of vehicles in vehicular Metaverses, thereby lacking in maintaining immersion. Subsequent research explores optimization strategies for resource allocation for VTs migration in vehicular Metaverses. The study in \cite{zhong2023blockchain} proposes a framework based on blockchain and game theory for optimizing resource allocation for VTs migration in vehicular networks. Additionally, \cite{wen2023task} introduces an incentive mechanism framework that utilizes the Age of Migration Task theory to efficiently allocate bandwidth resources for VTs migration in the vehicular Metaverses. 
Notably, there has been a lack of focus on applying auction theory to address the resource allocation problem during VTs migration. 

Unlike existing works that use game theory and contract theory to optimize resource allocation for VT migration in vehicular Metaverses, we apply auction theory to optimize resources for maximizing social welfare.
\begin{figure*}[t]
\centerline{\includegraphics[width=0.95\textwidth]{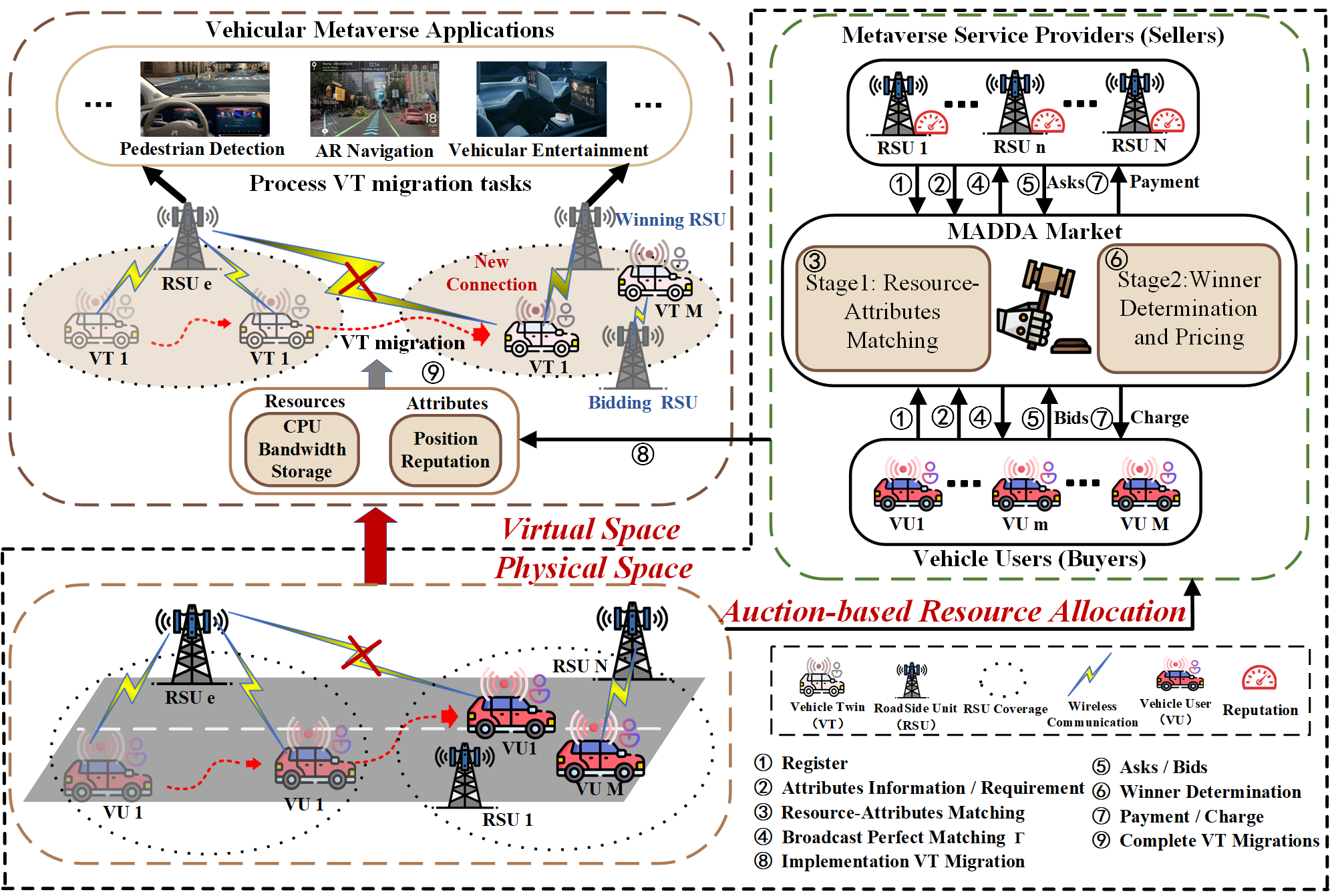}}
\caption{The muti-attribute double Dutch auction-based resource allocation mechanism for reliable VTs migration in vehicular Metaverses.}
\label{fig1}
\end{figure*}
\subsection{Auction for Resource Sharing}
In literature, auction-based resource trading in vehicular networks for making resource allocation and pricing decisions for limited resources in markets\cite{xu2023epvisa}. In \cite{10056728}, the authors introduce a truthful reverse auction that enables the VSP to select only IoT devices (e.g., images and videos) that enhance the quality of its virtual copy of objects through semantic information. However, this reverse auction is not well-suited for scenarios involving multiple sellers and multiple buyers. Consequently, the authors in \cite{9838736} developed a double Dutch auction for use in Wireless Edge-Empowered Metaverses, facilitating dynamic matching and pricing of VR services while ensuring incentive rationality, truthfulness, and budget balance. However, these auctions do not consider the heterogeneity of resources, making them less suitable for VTs migration systems with various VT tasks. Moreover, the above auctions for resource allocation are not suitable for maintaining the immersion of vehicular users during VTs migration since they merely consider the pricing factor in determining the winners. 

Unlike traditional auctions that primarily consider pricing factors, we propose a multi-attribute double Dutch auction to allocate and price VT migration tasks. We then apply a new GPT-based DRL algorithm, which is more suitable for solving sequence problem (e.g., the double Dutch clock sequence) to accelerate the auction process.


\section{System Model}\label{system}
In this section, we provide an overview of our system model for VTs migration in the vehicular Metaverses, which consists of two main parts, i.e., the reputation model to evaluate trust and reliability in vehicular networks\cite{zhong2023blockchain} and the valuation model to assess the economic impact and utility of VTs migration in vehicular Metaverses.  For convenience, we list the symbols used in our system model in Table~\ref{table I}.

Vehicular Metaverses are characterized by multiple vehicles in the physical space and their VTs in the virtual space, which are synchronized in real-time through radio access networks (RANs). In the physical space, we consider a set of $\mathcal{N}={\{1,\cdots, n,\cdots, N\}}$ comprising $N$ available RSUs with fixed positions $P_{n}=(x_n^t,y_n^t)$, functioning as MSPs, and a set $\mathcal{M}={\{1,\cdots, m,\cdots, M\}}$ of $M$ moving vehicles regarded as vehicular users (VUs). Importantly, both passengers and drivers in vehicles are the vehicular users, thereby enabling full immersion in services within the vehicular Metaverse, such as pedestrian detection on head-up displays, AR navigation, and VR entertainment, as depicted in Fig.~\ref{fig1}. In scenarios of resource-limited vehicles, the offloading of vehicular Metaverse services, i.e.,  VT tasks, occurs to nearby edge servers, e.g., RSUs, which possess significant resources $O_{n}=(O_{n}^{cp}, O_{n}^{com}, O_{n}^{s})$ for processing VT tasks, where $O_{n}^{cp}$, $O_{n}^{com}$, $O_{n}^{s}$ denote the amounts of computation, communication, and storage resources available to RSU $n$, respectively. 
As vehicle $m$ leaves the coverage of the current RSU at slot $t-1$, VU $m$ must enter the market to procure resources from new RSUs for VTs migration. Define $R_{m}=(R_{m}^{cp}, R_{m}^{com}, R_{m}^{s})$ as the required resources for VU $m$ to facilitate the VTs migration tasks, where $R_{m}^{cp}$, $R_{m}^{com}$, $R_{m}^{s}$ represent the computation, communication, and storage resources needed by VU $m$, respectively.

To achieve efficient VTs migration and maintain immersion for vehicular users, we propose a multi-attribute resource market for VTs migration to incentivize MSPs to serve VTs for vehicular users. This market not only considers the price attribute but also incorporates various non-monetary attributes in the auction process. We assume the existence of $L$ non-monetary attributes, which are represented as attributes set denoted by $\mathbf{A}=(a_1, \cdots, a_l, \cdots, a_L)$. Each VU exhibits unique weight requirements for these non-monetary attributes, tailored to their specific VTs migration tasks. Therefore, the multi-attribute weights of VU $m$ are captured by the weight vector $W_{m}=(\omega_{m1}, \omega_{m2}, \cdots, \omega_{mL})$. Consequently, the multi-attribute value requirement for the $L$ attributes by VU $m$ is represented as $Q_m=(q_{m1}, \cdots, q_{mi}, \cdots, q_{mL})$, where $q_{mi}$ signifies the expected attribute value of non-monetary attribute ${a_i}$ for VU $m$. In parallel, the $L$ attribute values held by MSP $n$ are denoted as $\hat{Q}_n=(\hat{q}_{n1}, \cdots, \hat{q}_{nj}, \cdots, \hat{q}_{nL})$, where $\hat{q}_{nj}$ indicates the attribute value of non-monetary attribute ${a_j}$ possessed by MSP $n$.

To contextualize our model, two critical attributes, i.e., location and reputation, are selected to exemplify their utility in enhancing immersive experiences for vehicular users. Due to the constrained coverage area of RSUs and the inherent mobility of vehicles, a single RSU is unable to provide uninterrupted services to vehicular users. Consequently, Location becomes a critical factor in determining the matching between vehicular users and RSUs in the VTs migration auction scenario. Before the VTs migration, it is considered that VU $m$ establishes a communication connection with RSU $e$ at the time slot $t-1$, the position of which is represented as $P_{m\rightarrow e}^{t-1}=(x_{e}^{t-1},y_{e}^{t-1})$. The distance $d_{e\to n}^t$ between the current RSU $e$ and the next RSU $n$ at time slot $t$ is defined as
 \begin{equation}
    d_{e\to n}^t=\sqrt{|x_{e}^{t-1}-x_n^t|^2+|y_{e}^{t-1}-y_n^t|^2}. \label{eq1}
\end{equation} Ensuring integrity in the performance of all RSUs within vehicular Metaverses presents significant challenges. For instance, the presence of malicious RSUs can compromise the immersive experience of vehicular users by manipulating location calculation results or impeding the execution speed. Consequently, the implementation of a robust reputation system is critical to guaranteeing reliable services. Therefore, we define the minimum multi-attribute requirements of VU $m$ as $Q_m=(q_{m1}, q_{m2})$, where $q_{m1}$ indicates the maximum distance VU $m$ is willing to migrate from the current RSU to the next RSU, and $q_{m2}$ represents the minimum reputation value sought by VU $m$ from the new RSU. Similarly, let $\hat{Q}_{n}=(\hat{q}_{n1},\hat{q}_{n2})$ indicate the multi-attribute value possessed by MSP $n$, where $\hat{q}_{n1}$ denotes the position of MSP $n$, i.e., $\hat{q}_{n1}= (x_{n}^{t},y_{n}^{t})$, and $\hat{q}_{n2}$ reflects the reputation value held by MSP $n$.

\subsection{Reputation Model}
In the VTs migration system, a reputation model is designed to compute the reputation value of MSP $n$ represented as $D_{n}$. $D_n$ falls within the range [0,1], with 0 indicating complete distrust and 1 indicating complete trust. The reputation value of MSP $n$ is influenced by the feedback evaluations from vehicular users who have had historical transactions with MSP $n$ and a time decay factor. Upon MSP $n$ accomplishing the VT migration task for VU $m$, VU $m$ reports to the auctioneer the actual quantity vector $S_{mn}$ of each type of resource provided by MSP $n$, as experienced during the execution of the VT migration task. Specifically, let $S_{mn}=(s_{mn}^{1}, \cdots,s_{mn}^{k},\cdots, s_{mn}^{K})$, where $s_{mn}^{k}$ denotes the quantity of the $k$-th resource actually provided by MSP $n$ to VU $m$. Subsequently, the auctioneer calculates the resource feedback evaluation of VU $m$ for the $k$-th types of resources offered by MSP $n$ after the transaction according to 
   \begin{equation}
   E_{mn}^k=\frac{\min{(R_m^k,s_{mn}^{k})}}{R_m^k}
   ,\label{eq10}
   \end{equation}
where $R_m^k$ is the quantity of the $k$-th type of resource required by VU $m$ according to $R_{m}$. For instance, there are three types of resources required for VU $m$ to facilitate VTs migration tasks, i.e., $R_{m}=(R_{m}^{cp}, R_{m}^{com}, R_{m}^{s})$. In this situation, the $1$-th type of resource is computation resources, i.e., $R_{m}^{1}\leftarrow R_{m}^{cp}$.

Due to the varied resource requests of vehicular users, each VU has specific resource needs, necessitating the consideration of resource weights on the resource feedback evaluation. Then, we let ${E}_{n}(a)$ denote the weighted feedback evaluation of all types of resources received by MSP $n$ in the $a$-th transaction, which can be calculated as 
   \begin{equation}
   E_{n}(a)=\sum_{k=1}^K\omega_{m}^k(a) E_{mn}^k
   ,\label{eq10}
   \end{equation}
   where $\omega_{m}^k(a)$ denotes the resource weight of VU $m$ for the $k$-th resource in the $a$-th transaction. 

  Since MSP $n$ exhibits varying reputation values in different transactions, to more accurately reflect the reputation value of MSP $n$, the reputation model also accounts for the diminishing impact of resource feedback evaluation over time. For updating the reputation value of MSP $n$ after the $\tau$-th transaction, the auctioneer must consider the resource feedback evaluations of $\tau$ transactions involving MSP $n$. We use $t_{n}(a)$ to indicate the transaction moment of MSP $n$ in the $a$-th transaction, satisfying $t_{n}(1)\leq \cdots \leq t_{n}(a)\leq \cdots \leq t_{n}(\tau)$. To quantify the effect of time decay on weighted feedback evaluations, $\lambda_{n}(a)$ denotes the freshness of weighted feedback evaluation $E_{n}(a)$ \cite{kang2019incentive},
  ensuring $\Sigma_{a=1}^\tau\lambda_{n}(a)=1$, where $\lambda_{n}(a)$ can be defined as
   \begin{equation}
   \lambda_{n}(a)=e^{-\xi (t_{\mu }-t_{n}(a))}
   ,\label{eq10}
   \end{equation}
   where $t_{\mu }$ is the moment when the auctioneer updates the reputation of MSP $n$, $\xi $ is the parameter determining  the rate of time decay, with the condition $\xi>0$.
   
   Considering the weighted feedback evaluation $E_{n}(a)$ as well as the freshness of $E_{n}(a)$, the reputation value of MSP $n$ can be calculated as 
   \begin{equation}
   D_{n}=\sum_{a=1}^\tau\lambda_{n}(a)E_{n}(a).\label{eq13}
   \end{equation}

\subsection{Valuation Model}
\subsubsection{\textbf{Design of value function}}
To facilitate VTs migration tasks effectively that meet the resource requirements $R_{m}=(R_{m}^{cp}, R_{m}^{com}, R_{m}^{s})$ of VM $m$, it is essential to account for the resources owned by MSPs. Typically, the value of computation resources is influenced by the CPU frequency of MSP $n$. Conversely, the value of communication resources fluctuates with the bandwidth, which is subject to time-varying demands. Likewise, the value of storage resources is based on the storage capacity of MSPs. Therefore, the MSPs determine their value functions $v_n^s$ based on their resource capacities, which can be represented specifically as 
\begin{equation}
v_n^s=\omega_1\delta_nf_n^2+\omega_2E_n \hat{B}_n+\omega_3x_n\varepsilon , \label{eq1}
\end{equation}
where $\omega_1$, $\omega_2$, and $\omega_3$ are weighting factors with the constraint $\omega_1+\omega_2+\omega_3=1$, $\delta_n$ represents the effective capacitance coefficient of MSP $n$'s computing chipset\cite{Burd1996ProcessorDF}, $f_n$ is the CPU frequency of MSP $n$'s computing chipset, $E_n$ indicates the spectrum efficiency of MSP $n$, $\hat{B}_n$ denotes its allocated bandwidth, $x_n$ indicates the storage capacity of MSP $n$, and $\varepsilon$ reflects the cost per TB of storage capacity.

In wireless communication\cite{xu2023reconfiguring}, the transmission rate provided to VU $m$ at the time slot $t$ can be estimated as 
\begin{equation}
\Pi_m^t=B_{m}\log_2\left(1+\frac{\rho h^0d^{-\epsilon }}{N_0}\right),
\label{eq2}
\end{equation}
where $B_{m}$ represents the bandwidth requested by VU $m$, $\rho$ is the average transmitter power of MSP $n$, $h^0$ represents the unit channel power gain, $d$ signifies the maximum tolerable distance of VU $m$ to migrate VT tasks from the current RSU to the next RSU, $\epsilon $ denotes the path-loss exponent, and $N_0$ is the average noise power.

The size of VTs migration tasks can be measured by the storage resources $R_{m}^{s}$ requested by VU $m$. Therefore, we can obtain the expected latency $\bar{T}$ of the VT migration task for VU $m$, depending on the transformation rate $\Pi_m^t$ and the size of the VT migration task $R_m^s$, which can be defined as
\begin{equation}
\bar{T}=\frac{R_m^s}{\Pi_m^t}
.\label{eq3}
\end{equation}
Therefore, the valuation function of VU $m$ for the requested resource $R_{m}$ is specifically manifested through the expected latency of VTs migration tasks, which is influenced by the quantity of the requested resource and can be expressed as
\begin{equation}
v_{m}^{b}=\hat{\alpha} \lg{\left(\frac{\hat{T}}{\bar{T}}\right) } 
,\label{eq4}
\end{equation}
where $\hat{T}$ is the maximum tolerable latency required by vehicular users to maintain immersion during VTs migration, and $\hat{\alpha}$ is a latency sensitivity factor that can be influenced by $R_{m}^{cp}$. The valuation model needs to ensure $\frac{\hat{T}}{\bar{T}}\geq 1$.

Combining the reputation model that prioritizes integrity with the valuation model that emphasizes efficiency helps to connect trustworthiness with resource optimization in our proposed VTs migration system.

\section{Multi-Attribute Double Dutch Auction}\label{multi}
The proposed MADDA mechanism includes two-stage matching for vehicular users and MSPs in the multi-attribute resource market. We will introduce these two stages and demonstrate that the proposed MADDA mechanism could satisfy the basic economic properties in this section.
\subsection{Market Design}
Considering multiple MSPs and VUs for VTs migration in the vehicular Metaverses, we utilize the MADDA mechanism for the VTs migration tasks in the multi-attribute resource market. The objectives of the MADDA mechanism are: (i) to determine the winner through a two-stage matching process; and (ii) to ascertain the purchase prices for the winning VUs and the sale prices for the winning MSPs. In the multi-attribute resource market for VTs migration in vehicular Metaverses, several entities perform distinct roles as follows.
\subsubsection{Vehicular users (Buyers)} In this market, each buyer, i.e., a VU requesting VT migration, compensates the seller to acquire VTs migration services. The buyers are required to submit their resources requirements denoted as $R_{m}=(R_{m}^{cp}, R_{m}^{com}, R_{m}^{s})$, minimum attributes requirements $Q_m=(q_{m1}, q_{m2})$, and attributes weights $W_{m}=(\omega_{m1},\omega_{m2})$ to the auctioneer when participating in the multi-attribute resource market. Only buyers that succeed the resource-attributes matching are allowed to submit their buy-bids to the auctioneer before the start of each round in the DDA. The buy-bid of buyer $m$ at round $t$ is evaluated as $g_{m}^{t}=v_{m}^{b}$, which denotes the maximum price that buyer $m$ is willing to pay for the VT migration task processed by the seller.
\subsubsection{Metaverse service providers (Sellers)} The sellers are the MSPs in the vehicular Metaverses that provide computation, communication, and storage resources for VUs to facilitate VTs migration. They are required to submit their owned resources, expressed as $O_{n}=(O_{n}^{cp}, O_{n}^{com}, O_{n}^{s})$ and owned attributes $\hat{Q}_{n}=(\hat{q}_{n1},\hat{q}_{n2})$ to the auctioneer when entering the multi-attribute resource market. Only sellers who meet the resource-attributes matching are allowed to submit their sell-bids to the auctioneer before the start of each round in the DDA. The sell-bid of seller $n$ in round $t$ is evaluated as $k_{n}^{t}=v_{n}^{s}$ representing the minimum price that the seller $n$ is willing to accept for providing resources during the VT migration of the buyer $m$.

\subsubsection{Auctioneer} In our model, the auctioneer performs two functions. Initially, the auctioneer broadcasts the resource-attributes perfect matching $\Gamma $, those buyers and sellers who can participate in the DDA, after executing the Kuhn-Munkres algorithm\cite{6084856}, with further elaboration provided in Section \ref{IV.B}. Subsequently, the auctioneer functions as a GPT-based DRL agent, aimed at learning an efficient auction policy. This is achieved by operating two Dutch clocks, one for buyers and one for sellers, with the goals of maximizing social welfare and reducing the cost of auction information exchange.

\subsection{Kuhn-Munkres Algorithm for MADDA Mechanism}\label{IV.B}
During the VTs migration, not all sellers can meet the buyer's requirements in terms of resources and multi-attribute. The first step is to establish a connection between buyer and seller based on these attributes and resources. A seller can connect with multiple buyers. The aim of our concept is therefore to bring sellers and buyers together in the best possible way and to give preference to those who best meet the buyer's requirements in terms of the resources and multi-attribute in order to increase the resource utilization.

In this work, we model the resource-attributes matching problem for buyers and sellers as a weighted bipartite graph to represent the matching process. The vertices represent the buyers and sellers, and the edges indicate the compatibility between the resource and multi-attribute requirements of the seller and the buyer. Each edge is assigned a weight, with a higher weight indicating a better match between a seller and a buyer. Therefore, our goal is to find a maximum weighted match in this graph, with the following steps:
\subsubsection{Submit Requirement / Information} 
When entering the multi-attribute resource market, VU $m$ is required to submit the resources requirements $R_{m}=(R_{m}^{cp}, R_{m}^{com}, R_{m}^{s})$ and the minimum multi-attribute requirements $Q_m=(q_{m1},q_{m2})$ to the auctioneer. Simultaneously, MSP $n$ submits available resources information $O_{n}=(O_{n}^{cp}, O_{n}^{com}, O_{n}^{s})$ with the multi-attribute $\hat{Q}_n=(\hat{q}_{n1},\hat{q}_{n2})$ to the auctioneer.
\subsubsection{Establish Connection Based on Muti-attribute and Resources}
We define a matching between VU $m$ and MSP $n$ as existing 
(i.e., $m\leftrightarrow n$) if the available resources of MSP $n$ are greater than the resource requirements of VU $m$, and the multi-attribute profile of MSP $n$ satisfies the multi-attribute requirements of VU $n$. Therefore, the aforementioned constraints can be formulated as
\begin{equation}
\begin{aligned}
&(O_{n}^{cp} \ge R_{m}^{cp} )\cap (O_{n}^{cpm} \ge  
R_{m}^{com} )\cap (O_{n}^{s} \ge R_{m}^{s})\cap\\& (d_{e\to n}^{t}\le q_{m1} )\cap (\hat{q}_{n2}\ge q_{m2})
.\label{eq10}
\end{aligned}
\end{equation}

Based on the matching between the VU and MSP, we can construct an unweighted bipartite graph $G=(\mathcal{M}\cup \mathcal{N},\mathcal{M}\leftrightarrow \mathcal{N})$\cite{8952808}, where $\mathcal{N}$ and $\mathcal{M}$ are two sets of MSPs and VUs, denoting $N$ MSPs and $M$ VUs respectively in the multi-attribute resource market. $\mathcal{M}\leftrightarrow \mathcal{N}$ indicates the matches between MSPs and VUs.                                                                      
\subsubsection{Setting the Weight of Each Edge}
To accomplish our goal of assigning additional resources and greater attributes to buyers with greater resource and multi-attribute needs, we define the weights as the result of multiplying the buyer's requirements by the seller's resources. Let $\Upsilon(m,n)$ represent the weight of an edge $(m\leftrightarrow n)$ and the coverage of each RSU is $d_{c}$, which can be defined as 
\begin{equation}
\begin{aligned}
\Upsilon(m,n)= \omega_{m1}(d_{c}-d_{e\to n}^{t})+\omega_{m2}~\hat{q}_{n2}+R_{m}\cdot O_{n}
.\label{eq10}
\end{aligned}
\end{equation}
Then, by assigning the weights to each match, we can obtain the weighted bipartite graph $\hat{G} =(\mathcal{M}\cup \mathcal{N},\mathcal{M}\leftrightarrow \mathcal{N})$.
\subsubsection{Resource-attributes Matching of Weighted Bipartite Graph}
We find the maximum weighted matching of the weighted bipartite graph to achieve the goal mentioned in (3) of allocating more resources and larger attributes to buyers with greater resource and multi-attribute needs.
To find the maximum weighted match from the weighted bipartite graph $\hat{G} =(\mathcal{M}\cup \mathcal{N},\mathcal{M}\leftrightarrow \mathcal{N})$, we use the Kuhn-Munkres algorithm to solve this problem. The detailed steps are described in Algorithm \ref{A1}. First, to create a balanced bipartite graph $\hat{G}^{'}$, we introduce virtual vertices and edges with a weight of  ``$-1$'' into $\hat{G}$.
Next, we initialize flag $l(\cdot)$ of every vertex and construct an equality subgraph $\hat{G}_{l}^{'}$ where the flags are adjusted during the matching process and the $\hat{G}_{l}^{'}$ satisfies the equation in line 7 of Algorithm \ref{A1}.
\begin{equation}
\hat{G}_{l}^{'}=\{(m,n)\Rightarrow l(m)+l(n)=\Upsilon(m,n)\}
.\label{eq10}
\end{equation}
Finally, we choose the Hungarian algorithm\cite{mills2007dynamic} to achieve the perfect matching. If a perfect matching is found, the algorithm removes virtual vertices and their associated edges, resulting in the resource-attributes perfect matching $\Gamma $. If not, the algorithm iteratively refines the flags and continues the search, adapting the flags and repeating the process (lines $9-18$ ) until the resource-attributes perfect matching is achieved. For every edge $(m\leftrightarrow n)$ in $\Gamma $, it means that MSP $n$ can handle the VT migration task for VU $m$ based on the resources and multi-attribute requirements of VU $m$.

By following the above steps, we obtain the resource-attributes perfect matching $\Gamma $ of the weighted bipartite graph $\hat{G}$. Only the buyers and sellers within $\Gamma $ are eligible to submit buy-bids and sell-bids.

\begin{algorithm}[t]
\caption{Resource-attributes Matching Algorithm}
\label{A1}
\KwIn{${\hat{G}} =(\mathcal{M}\cup \mathcal{N},\mathcal{M}\leftrightarrow \mathcal{N})$}
\KwOut{resource-attributes perfect matching $\Gamma $  }
To construct a balanced bipartite graph $\hat{G}^{'}$, incorporate virtual vertices and edges with a weight of -1 into $\hat{G}$.

\textbf{Initialization:} Setting the starting flag $l(\cdot)$ for each vertex in $\hat{G}^{'}$.

\For{each $m\in \mathcal{M}$}{$l(m)$ is set to $ max\{n\in \mathcal{N}|\Upsilon(m,n)\}$};
\For{each $n\in \mathcal{N}$}{$l(n)$ is set to $ 0$}
Build equality subgraph $\hat{G}_{l}^{'}$ that satisfies
$\hat{G}_{l}^{'}=\{(m,n)\Rightarrow l(m)+l(n)=\Upsilon(m,n)\}$;

Selecting an arbitrary matching $\Gamma $ within $\hat{G}_{l}^{'}$;\\
Create sets $L$ and $R$ to store unsaturated and saturation points of $\Gamma $ when the Hungarian algorithm finishes;\\
Obtain $(Result, L, R)$ using Hungarian Algorithm$(\Gamma,\hat{G}_{l}^{'})$;

\If{$Result ~ is ~ False$}
{Calculate the minimum labeling $\delta _{l}$;
\\Update labels for $L$ and $R$ based on $\delta _{l}$;\\
Update the equality subgraph $\hat{G}_{l}^{'}$ accordingly;\\Go to line 9;}
\If{$Result ~ is ~ True$}
{Eliminate the virtual vertices and their associated edges from $\Gamma$.}
\end{algorithm}

\subsection{Auction Design}
In this section, we introduce the Double Dutch Auction (DDA) mechanism ~\cite{9838736}, which has been used for efficient resource allocation by combining the features of both dutch auctions and double auctions. In typical Dutch auctions ~\cite{bagwell1992dutch}, the auctioneer begins with a high asking price, which is lowered incrementally until a participant accepts the current price. Conversely, in double auctions ~\cite{mcafee1992dominant}, both buyers and sellers submit buy-bids and sell-bids, and transactions occur when compatible buy-bids and sell-bids meet. In DDA, buyers' bids decrease from high to low, while sellers' asks increase from low to high. The auction continues with these adjustments until a buy-bid and a sell-bid intersect, at which point the auction ends, and market clearing occurs. This mechanism ensures an efficient allocation of resources by dynamically adjusting prices to reach an equilibrium that satisfies both buyers and sellers. Then, we introduce the DDA mechanism in detail.
\begin{table*}[]
\centering
\caption{The contributions of algorithm design compared with current references.}
\begin{tabular}{|c|m{4cm}|m{4cm}|m{7cm}|}
\hline
         \textbf{Reference} & \centering\textbf{Algorithm Type} & \centering\textbf{Auction Mechanism} & \centering\textbf{Contributions}\tabularnewline
         \hline
         Our paper & Fusion of Weighted Bipartite Graph for resource-attribute matching and GPT-based Deep Reinforcement Learning (DRL) & Double Dutch Auction with multiple attributes such as price, location, and reputation & Combines the KM algorithm for resource-attribute matching with the GPT-based DRL algorithm to solve the complex problem of allocating and pricing multi-attribute resources in a double Dutch auction. \\ \hline
         \cite{6084856} & Kuhn-Munkres (KM) Algorithm; Bipartite Graph Matching & No auction mechanism & Utilizes a classic graph matching algorithm without considering dynamic auction-based resource allocation.\\ \hline
         \cite{9838736} & Basic DRL; No matching algorithm & Double Dutch Auction with price attribute & Applies basic DRL algorithm to optimize auction strategy, but does not include multi-attribute considerations.\\ \hline
          \cite{mills2007dynamic} & Hungarian Algorithm Matching; No reinforcement learning & No auction mechanism & Implements a Hungarian algorithm for resource matching without reinforcement learning or auction-based mechanisms, focusing on static optimization techniques.\\ \hline

\end{tabular}
\label{tab:my-table}
\end{table*} 


In DDA, the auctioneer initializes and adjusts two Dutch clocks, i.e., $C_{B}$ and $C_{S}$, for the buyer side and the seller side, respectively. $C_{B}$ represents the purchase price for each round and serves as the buyer's clock, while $C_{S}$ represents the sale price for each round and serves as the seller's clock. The buyer's clock and the seller's clock, denoted by $C_{B}$ and $C_{S}$, respectively, operate alternately in the DDA system. The active side of the auction clock is indicated by $\Psi^{t}$, where $\Psi^{t}=0$ for the buyer's clock and $\Psi^{t}=1$ for the seller's clock. The buyer's clock $C_{B}$ is initialized with the highest price $C_{B}^{0}=p^{\text{max}}$ and decreases with each round of the auction. Conversely, the seller's clock $C_{S}$ is initialized with the lowest price $C_{S}^{0}=p^{\text{min}}$ and increases with each round of the auction.

Assuming that the auctioneer uses the Kuhn-Munkres algorithm, the resulting number of optimally matched pairs of resource attributes in $\Gamma$ is denoted by $k$. This indicates that there are $k$ participating buyers and sellers in the bidding process. The DDA auction initiates with the buyers' round. At the auction's commencement, the $k$ buyers are ranked in a non-ascending sequence according to their values, denoted as $B=\{m|\forall m\in [1,k)\Rightarrow v_{m}^{b}\geq v_{m+1}^{b}\}$. Similarly, the $k$ sellers are sequenced in non-decreasing order based on their values, denoted as $S=\{n|\forall n\in [1,k)\Rightarrow v_{n}^{s}\leq v_{n+1}^{s}\}$. The DDA comprises the following five phases: Dutch Clock Initialization, Dutch Clock Broadcast, Dutch Clock Acceptance, Dutch Clock Adjustment, and Dutch Clock Termination.

\subsubsection{Dutch Clock Initialization}
At the beginning of the auction, the buyer clock $C_{B}$ is set at its highest price $C_{B}^{0}=p^\emph{max}$ and the seller clock denoted as $C_{S}$ is set at its lowest price $C_{S}^{0}=p^{min}$. Meanwhile, all buyers and sellers have not yet agreed to the clock price, meaning both the buy-winner sequence $M_{B}^{t} $ and the sell-winner sequence $N_{S}^{t} $ are initially empty. In addition to these sequences, two sequences $H_{B}$ and $H_{S}$ are used to log the received price of the two Dutch clocks. At the start of the auction, $H_{B}^{0}=\emptyset$ and $H_{S}^{0}=\emptyset$.

\subsubsection{Dutch Clock Broadcast} The auctioneer announces the buyer clock $C_{B}^{t}$ in the market when $\Psi^{t}=0$, and announces the seller clock $C_{S}^{t}$ when $\Psi^{t}=1$.

\subsubsection{Dutch Clock Acceptance} Upon receiving the auction clock announcement by the auctioneer, the buyers or sellers evaluate whether to accept the current clock by comparing it with their bids. During the buyers' phase, i.e., $\Psi^{t}=0$, the $m$-th buyer, who is yet to bid, will accept the current clock $C_{B}^{t}$ as their buy-bid $g_{m}^{t}$, if $C_{B}^{t}$ is lower than their expected maximum buy-bid $v_{m}^{b}$, i.e., $g_{m}^{t}\leftarrow C_{B}^{t} \Leftarrow C_{B}^{t}\le v_{m}^{b}$. Subsequently, buyer $m$ will be included in the buy-winner sequence, i.e., $M_{B}^{t+1}= M_{B}^{t}\cup {m}$. Hence, the difference between the expected buy-bid and the actual buy-bid constitutes the regret $\phi _{m}^{t}$ of buyer $m$, which can be represented as
\begin{equation}
\phi _{m}^{t}=v_{m}^{b}-g_{m}^{t}
.\label{eq10}
\end{equation}
During the sellers' phase, i.e., $\Psi^{t}=1$, the $n$-th seller, who is yet to place a sell-bid, will accept the current clock $C_{S}^{t}$ as their sell-bid $k_{n}^{t}$ provided $C_{S}^{t}$ exceeds their expected minimum sell-bid $v_{n}^{s}$, i.e., $k_{n}^{t}\leftarrow C_{S}^{t} \Leftarrow C_{S}^{t}\ge v_{n}^{s}$. Subsequently, the auctioneer incorporates seller $n$ into the sell-winner sequence $N_{S}^{t}$, i.e., $N_{S}^{t+1}= N_{S}^{t}\cup {n}$. Consequently, the disparity between the anticipated sell-bid and the actual sell-bid constitutes the regret $\phi _{n}^{t}$ for seller $n$, which can be expressed as
\begin{equation}
\phi _{n}^{t}=k_{n}^{t}-v_{n}^{s}
.\label{eq10}
\end{equation}

\subsubsection{Dutch Clock Adjustment}
In the buyer round, i.e., $\Psi^{t}=0$, should no buyer agree to the current clock, the auctioneer must select a step size that corresponds to a multiple of the minimum price interval $\alpha^{t}$, to reduce the price of the buyer clock, i.e., $C_{B}^{t+1} \Leftarrow C_{B}^{t} - \alpha^{t}$. Conversely, during the seller round, i.e., $\Psi^{t}=1$, if no seller agrees to the current seller clock, the auctioneer will raise the price of the seller clock, i.e., $C_{S}^{t+1} \Leftarrow C_{S}^{t} + \alpha^{t}$.

\subsubsection{Dutch Clock termination}

Following each adjustment of the auction clock, the auctioneer verifies whether the two Dutch clocks have crossed. When the two Dutch clocks crossed, i.e., $C_{B}^{t} < C_{S}^{t}$, the auction concludes at time $T\leftarrow t$. Besides, the market clearing price is determined as the average of the two Dutch clocks, i.e., $p^{\ast} = \alpha C_{B}^{T-1} + (1-\alpha) C_{S}^{T-1}$, where $\alpha \in [0,1]$ represents the price effectiveness factor.


Let $\Lambda \Leftarrow\{\forall m\in M_{B}^{t},~ \forall n\in N_{S}^{t}| \exists~ (m\leftrightarrow n)\in \Gamma \}$ denote the set of candidate winning pairs which comply with the condition that the candidate winning pairs are included in the resource-attributes perfect matching $\Gamma$, i.e., $\Lambda \subseteq \Gamma$. In scenarios where the market is cleared when $\Psi^{t}=1$, the first $|\Lambda|-1$ candidate winning pairs in $\Lambda$ constitute the ultimate winning pairs, with $|\Lambda|$ representing the number of candidate winning pairs in $\Lambda$. Alternatively, if the market is cleared when $\Psi^{t}=0$, the first $|\Lambda|$ candidate winning pairs emerge as the auction winners. Social welfare ($SW$) is the aggregate of the ultimate winners' utility, comprising both the seller utility $U_{S}$ and the buyer utility $U_{B}$, i.e., $SW=U_{S}+U_{B}$. The utility of buyer $m$ is expressed as $u_{m}$, while the utility of seller $n$ is articulated as $\hat{u}_{n}$ for those instances where buyer $m$ and seller $n$ form part of the ultimate winning pairs $\hat{\Lambda}$, which can be calculated as 
\begin{equation}
u_{m}=\begin{cases}
  g_{m}^{t}-p^{\ast}, & \text{ if } m\in \hat{\Lambda},   \\
  ~~~~0,& \text {otherwise},
\end{cases}
\label{eq10}
\end{equation}
and
\begin{equation}
\hat{u}_{n}=\begin{cases}
  p^{\ast }-k_{n}^{t}, & \text{ if } n\in \hat{\Lambda},  \\
  ~~~~0,& \text {otherwise}.
\end{cases}
\label{eq10}
\end{equation}
Therefore, $U_{B}= {\textstyle \sum_{m=1}^{\kappa }} g_{m}^{t}-p^{\ast } $ and $U_{S}= {\textstyle \sum_{n=1}^{\kappa }} p^{\ast }- k_{n}^{t}$, where $\kappa $ denotes the number of ultimate winning pairs $\hat{\Lambda } $.






\subsection{Economic Properties}

In designing an auction for VT migration task requests based on auctions, our objective is to attain the following economic properties. First, the auction should ensure a balanced budget to maintain the market's sustainability, indicating that the financial exchanges among participants are balanced. Second, the auction should uphold Individual Rationality (IR) and Incentive Compatibility (IC), signifying that participants need to gain non-negative utility and are discouraged from benefitting through unreal bid submissions.
Let $c=(c_{1},\cdots,c_{m},\cdots,c_{\kappa})$ represent the charge imposed by the auctioneer on the buy-winner and $p=(p_{1},\cdots,p_{n},\cdots,p_{\kappa})$ represent the payment made by the auctioneer to the sell-winner. Subsequently, IR, IC, and Budget Balancing are elaborated as follows.

\begin{definition}[Individual Rationality]
    Individual rationality ensures that each winning buyer and seller secures non-negative utility in an auction, i.e., $g_{m}^{t}\ge c_{m}$ for the buyers and $k_{n}^{t}\le p_{n}$ for the sellers.
\end{definition}



\begin{definition}[Incentive Compatibility]
     Asserts that buyers and sellers are reluctant to enhance their utility by submitting bids or asks that deviate from their genuine valuation, i.e., $g_{m}^{t} \ne v_{m}^{b}$ for buyers or $k_{n}^{t} \ne v_{n}^{s}$ for sellers.
\end{definition}

\begin{definition}[Strong Budget Balance]
    When the auctioneer achieves non-negative utility, i.e., the difference between total costs and total payments is positive, the double auction is deemed to uphold a budget balance. Furthermore, auctions exhibit a strong budget balance when no payment transfers or costs are incurred by the auctioneer.
\end{definition}

In alignment with the conventions in existing literature, auctions are deemed economically robust if they embody IR, IC, and budget balanced principles. Initially, it was demonstrated that MADDA upholds a strong budget balance.

\begin{theorem}\label{T1}
    The proposed MADDA upholds a strong budget balance.
\end{theorem}

\begin{proof}
    According to $\Lambda \Leftarrow\{\forall m\in M_{B}^{t},~ \forall n\in N_{S}^{t}| \exists~ (m\leftrightarrow n)\in \Gamma \}$, we can obtain the candidate winning pairs $\Lambda$. And the ultimate winning pairs depend on the active side $\Psi^{t}$ when the market is cleared. In scenarios where the market is cleared when $\Psi^{t}=1$, the first $|\Lambda|-1$ candidate winning pairs in $\Lambda$ constitute the ultimate winning pairs. Alternatively, if the market is cleared when $\Psi^{t}=0$, the first $|\Lambda|$ candidate winning pairs emerge as the auction winners. Consequently, in the MADDA mechanism, the number of winning buyers equals the number of winning sellers after the auction ends. Additionally, the winning buyers and the winning sellers transact at the market clearing price $p^{\ast}$ as the trading price. Therefore, for each ultimate winning pair, the auctioneer's utility is always 0. Thus, the aggregate utility of the auctioneer totals 0, fulfilling the criteria for a strong budget balance.
\end{proof}

Following the proof in \cite{6330963}, we demonstrate that MADDA is doubly monotone and critical, opting not to offer direct proof for IR and IC. Based on these two properties, then, we use Theorem \ref{T2} to prove the MADDA is IR and IC. Monotonicity and criticality are defined as follows.

\begin{definition}[Monotonicity]
    An auction mechanism is bid monotonic such that for each buyer $m$ (seller $n$), if it wins the auction by submitting $g_{m}^{t}$ ($k_{n}^{t}$), and then it wins by submitting $g_{m}^{'} > g_{m}^{t}$ ($k_{n}^{'} < k_{n}^{t}$), given the other's submission which remain unchanged.
    
\end{definition}

\begin{definition}[Criticality]
    For a winning buyer $m$, the price $p^{\ast }$ is deemed critical if buyer $m$ secures a win by submitting $g_{m}^{t} > p^{\ast}$ and incurs a loss with a submission of $g_{m}^{t} < p^{\ast}$, assuming the other participants' bids do not vary.
\end{definition}


\begin{theorem}\label{T2}
    A bid-monotonic auction is IR and IC if and only if there is a critical charge to buyers and a critical payment to sellers.
\end{theorem}

\begin{lemma}\label{L1}
    The proposed MADDA is bid monotonic.
\end{lemma}

\begin{proof}
    Suppose a buyer $m$ in $M_{B}^{t}$ has a current bid of $g_{m}^{t}$. If the buyer opts to increase their bid to $g_{m}^{'} > g_{m}^{t}$, this indicates that buyer $m$ is willing to accept a higher buyer clock for the same set of resources and multi-attribute offered. Given that the candidate winning pairs $\Lambda$ rely on the resource-attributes perfect matching $\Gamma$, which is pre-established, the enhanced bid of buyer $m$ does not impact their eligibility for being matched. Hence, if buyer $m$ was part of a candidate winning pair in $\Lambda$ before the bid increases, they will remain in $\Lambda$ after the bid increases. This preserves their chance of being among the ultimate winners $\hat{\Lambda}$, provided $\Psi^{t}$ conditions are satisfied.

In a parallel scenario, for a seller $n$ in $N_{S}^{t}$ with a current bid of $k_{n}^{t}$, lowering their bid suggests a willingness to accept a lower seller clock for their resources and multi-attribute. Echoing the buyer's case, the decreased bid does not influence the seller's inclusion in the candidate winning pairs $\Lambda$, as the matching is anchored in $\Gamma$. Consequently, if seller $n$ was part of a candidate winning pair before their bid decrease, they will remain in $\Lambda$ after the bid decreases, improving their opportunity among the ultimate winners.

The market clearing condition $\Psi^{t}$ dictates the final tally of winning pairs. As this condition is contingent upon individual bids, it does not compromise the bid monotonicity principle.
\end{proof}


According to Theorem \ref{T2}, to prove that the proposed MADDA satisfies the IR and IC principles, in addition to proving that the MADDA is bid monotonic, we also need to prove that there is a critical payment for all winning buyers and winning sellers.

 \begin{lemma}\label{L2}
     In the proposed MADDA, $p^{\ast }$ is the critical payment for all winning buyers and winning sellers.
 \end{lemma}

\begin{proof}
    In MADDA, buyers and sellers are matched based on multi-attribute criteria. The resource-attributes perfect matching $\Gamma$ generally takes place prior to the auction's commencement, indicating that resource-attributes matching is an independent step that precedes the auction process. Despite the implementation of the resource-attributes matching, the pricing process remains consistent with the principles of the DDA.

When $\Psi^{t}=0$, the auctioneer determines the clearing price considering both the buyer's and seller's clocks. On the buyer side, characterized by clearing through trade reduction, the buyers and sellers involved in the calculation of this clearing price do not necessarily become winning buyers or winning sellers. As a result, both the winning buyers and sellers, as well as the remaining sellers in the market, do not affect the market clearing price $p^{\ast}$.

On the contrary, if the auction ends in the seller side, the market clearing price is established based on the buyer and seller clocks before any adjustments are made by the auctioneer. In this scenario, the final buyer influences the formation of the price but does not engage in the auction. This implies that although the winning sellers and buyers in the market may adjust their bids, these changes do not affect the market clearing price.

Thus, the market clearing price plays a pivotal role in both seller’s and buyer’s markets. The resource-attributes perfect matching $\Gamma$ serves as a step in determining the matching relationship between buyers and sellers and does not alter the pricing rules defined by the DDA mechanism. Therefore, $p^{\ast}$, as the critical transaction value for all winning buyers and sellers, holds validity in MADDA.
\end{proof}

However, demonstrating that MADDA upholds economic robustness as specified in Theorems \ref{T1} and \ref{T2} necessitates the premise that the resource-attributes matching process is stable, i.e., the resource-attributes matching process operates independently from the auction process.

Following the process in \cite{chegireddy1987algorithms}, we establish that the resource-attributes matching process maintains stability using Theorem \ref{T3}. We initially present the definition of feasible labels.

\begin{definition}[Feasible Labels]
    Feasible labels for a bipartite graph $G$ entail assigning a weight $l(\cdot)$ to each vertex $m$ ($n$) in a manner that, for all edges $(m\leftrightarrow n)$ within the graph, the condition $\Upsilon(m,n) \le l(m) + l(n)$ is satisfied. These collections of vertex weights constitute the feasible labels for the graph.
\end{definition}

With the definition of feasible labels, we can obtain the maximum weight perfect matching of the bipartite graph according to Lemma \ref{L3}.

\begin{lemma}\label{L3}
    If a bipartite graph $G$ possesses a set of feasible vertex labels, and if the equality subgraph $\hat{G}_{l}^{'}$ associated with this feasible vertex label exhibits a resource-attributes perfect matching $\Gamma$, then this matching is the maximum weight perfect matching of the original bipartite graph, as detailed in \cite{chegireddy1987algorithms}.
\end{lemma}
\begin{proof}
    Consider any perfect matching $\hat{M}$ of the original bipartite graph, where the sum of the weights of the edges in $\hat{M}$, referred to as $val(\hat{M})$, is the cumulative weight of the matched edges (matched edges which do not share common vertices). According to the definition of feasible labels, it can be inferred that the sum of the weights of the edges in any perfect matching is at most equal to the sum of the weights of the vertices in any feasible labels, i.e., $val(\hat{M})=\sum_{(m,n)}^{\hat{M}} \Upsilon (m,n) \le \sum_{i=1}^{a} l(i)$, where $a$ represents the total number of feasible labels.
When there exists a set of feasible labels and the equality subgraph $\hat{G}_{l}^{'}$ of these labels exhibits a resource-attributes perfect matching, the sum of the weights of the edges in this resource-attributes perfect matching of the equality subgraph, $\Gamma$, termed $val(\Gamma)$, is calculated as $val(\Gamma)=\sum_{(m,n)}^{\Gamma}\Upsilon (m,n) =\sum_{(m,n)}^{\Gamma} l(m)+l(n)=\sum_{i=1}^{a}l(i)$. It is evident that for any perfect matching $\hat{M}$, $val(\hat{M}) \le val(\Gamma)$. Therefore, $\Gamma$ constitutes the resource-attributes perfect matching with the maximum sum of weights, thereby qualifying as the resource-attributes perfect matching.
\end{proof}

\begin{table*}[]
\centering
\caption{The contributions of economic properties compared with current references.}
\begin{tabular}
{|c|>{\centering\arraybackslash}m{5cm}|>{\centering\arraybackslash}m{2cm}|>{\centering\arraybackslash}m{2cm}|>{\centering\arraybackslash}m{2cm}|>{\centering\arraybackslash}m{2cm}|}
\hline
\textbf{Reference} & \textbf{Auction Mechanism} & \textbf{Individual Rationality} & \textbf{Budget Balance} & \textbf{Incentive Compatibility} & \textbf{Social Welfare} \tabularnewline
\hline
Our paper & Multi-attribute Double Dutch Auction & \cmark & \cmark & \cmark & \cmark \\ \hline
\cite{8952808} & Multi-attribute Double Auction & \cmark & \cmark & \cmark & \xmark \\ \hline
\cite{6330963} & Spectrum Double Auction & \cmark & \cmark & \cmark & \xmark \\ \hline
\cite{chegireddy1987algorithms} & Just a k-best Perfect Matching Algorithm & \xmark & \xmark & \xmark & \xmark \\ \hline
\end{tabular}
\label{tab:my-table}
\end{table*}
\begin{theorem}\label{T3}
    The resource-attributes matching is independent of the MADDA's pricing and allocation processes.
\end{theorem}
\begin{proof}
It can be proved by Lemma \ref{L3} that the resource-attributes perfect matching $\Gamma$ is the maximum weight perfect matching of the equality subgraph $\hat{G}_{l}^{'}$. Meanwhile, it can also be concluded from Algorithm \ref{A1} that the resource-attributes perfect matching $\Gamma$ is the only one \cite{8952808} and is only related to the situation of supply and demand in the multi-attribute resource market, i.e., the resource-attributes demand of buyers and the resource-attributes supply of sellers. Another, in the proposed MADDA mechanism, only the buyers and sellers in the resource-attributes perfect matching $\Gamma$ can participate in the DDA to submit their buy-bids and sell-bids but no change the connections between the buyers and the sellers in the resource-attributes perfect matching $\Gamma$. This implies that regardless of how the auction process evolves, the outcomes of the resource-attributes perfect matching $\Gamma$ remain constant, and the resource-attributes matching process itself is independent and unaffected by dynamic changes in the DDA. Therefore, the resource-attributes matching process maintains stability.
    
\end{proof}

\begin{figure}
    \centering
    \includegraphics[width=1\linewidth]{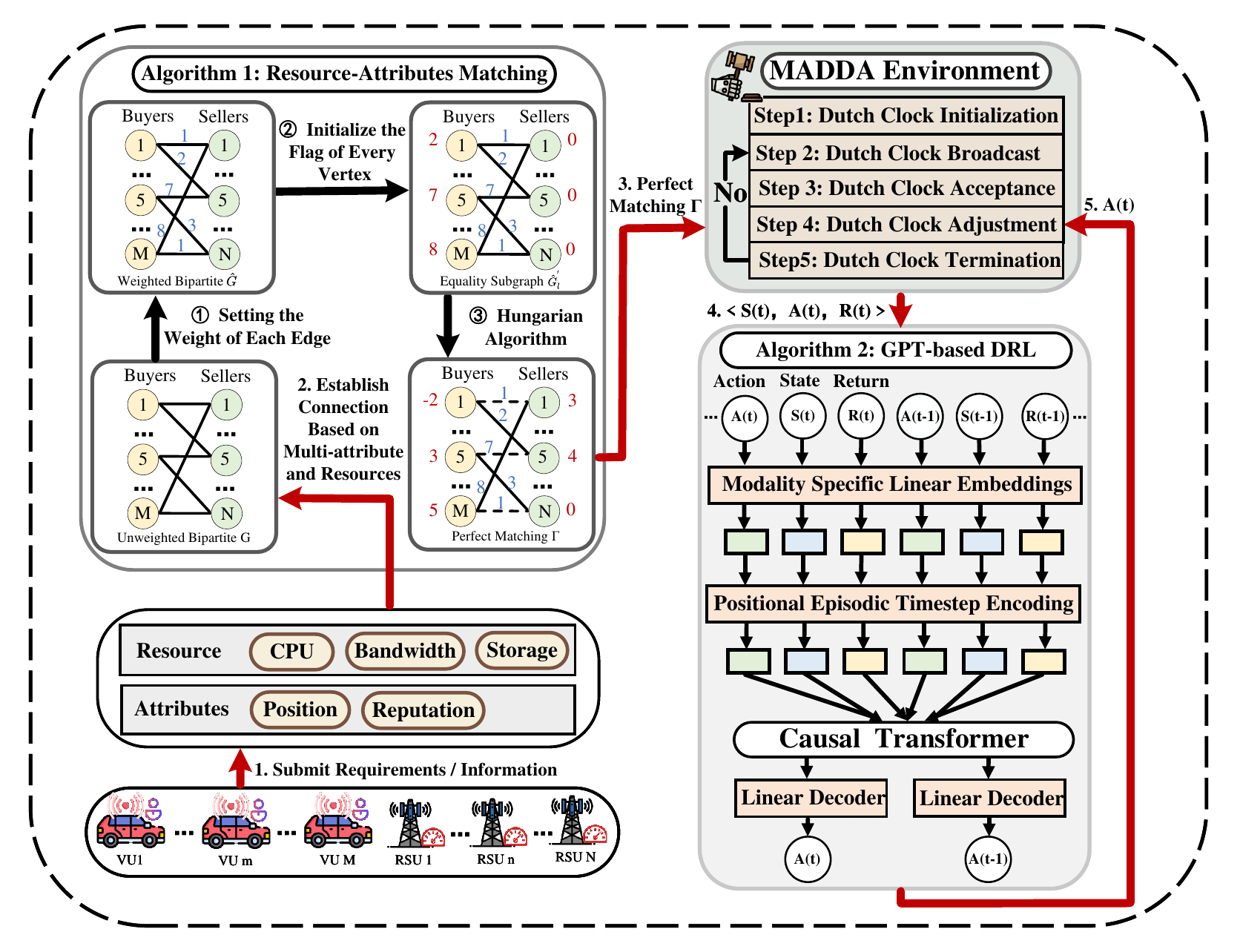}
    \caption{The proposed efficient GPT-based incentive mechanism to improve the efficiency of the DDA. }
    \label{fig: GPT-based DRL}
\end{figure}

\section{GPT-based Deep Reinforcement Learning for MADDA}\label{V}
To address the economic challenge of maximizing social welfare while maintaining IR, ensuring IC, and achieving budget balance in the MADDA, we transform the MADDA process into a Markov Decision Process. In this process, the auctioneer, acting as a DRL agent, interacts with the environment, without the need for priori auction knowledge.

\subsection{Markov Decision Process for MADDA}
A Markov Decision Process is represented by $<S, A, P, R > $, where $S$ denotes the state space, $A$ indicates the action space, $P$ refers to the probability of state transitions, and $R$ refers to the reward. Then, we describe the state space, action space, and reward in detail.



\subsubsection{State Space}  In each clock adjustment round $t$ $(t=1,2,\cdots, T)$, the state space encompasses several elements: the auction flag $\Psi^{t}$, the current auction round number $t$, two Dutch clocks $C_{B}^{t}$ and $C_{S}^{t}$, the numbers of the buy-winner sequence $|M_{B}^{t}|$ and the sell-winner sequence $|N_{S}^{t}|$, collectively represented as $S^{t} \triangleq \{\Psi^{t}, t, C_{B}^{t}, C_{S}^{t}, |M_{B}^{t}|, |N_{S}^{t}|\}$.
\subsubsection{Action Space}During the clock adjustment, the auctioneer alternates between the buyer clock and the seller clock with an available step size $\alpha^{t}$ at decision slot $t$, i.e., $\forall t,\alpha^{t}\in A$.
\subsubsection{Reward} In MADDA, the reward of the auctioneer stems from two key factors: the trade regret experienced by both buyers and sellers and the cost incurred in broadcasting new auction clock information to either buyers or sellers, which can be defined as
\begin{equation}
    r(S^{t},S^{t+1})=\begin{cases}
  & -\phi _{m}^{t},~~~~~\text{ if } \Psi ^{t}=
  0, g_{m}^{t}\ge C_{B}^{t},\\
  & -{c}_{t}(k),~~\text{ if } \Psi ^{t}=0,g_{m}^{t}\le  C_{B}^{t},\\
  & -\phi _{n}^{t},~~~~~\text{ if }\Psi ^{t}= 1,k_{n}^{t}\le  C_{S}^{t},  \\
  & -{c}_{t}(\hat{k}),~~\text{ if }\Psi ^{t}= 1,k_{n}^{t}\ge  C_{S}^{t}, 
\end{cases}
\end{equation}
where $c_{t}(X)$ denotes the function that calculates the cost of exchanging auction information for $X$ participants in a market of this size. This function is defined as $c_{t}(X)=\zeta X$, with $\zeta$ symbolizing the penalty associated with communication during the auction.

The reward function's design aims to encourage the auctioneer to minimize trade regret and maximize matching pairs to enhance the social welfare (SW) in fewer auction rounds. The reward function comprises two components: (i) a trade regret term ($\phi _{m}^{t}$ or $\phi _{n}^{t}$) to reflect the authenticity of the auction, (ii) a negative communication cost term $-c_{t}(X)$ to diminish the number of participants requiring new auction clock updates, thus reducing communication costs\cite{9838736}. These considerations incentivize the auctioneer to find an optimal step size for adjusting the Dutch clock, striking a balance between accelerating the auction process and ensuring that potential buyers and sellers are given adequate bidding opportunities.

\begin{algorithm}
\caption{The GPT-based DRL algorithm}\label{alg:decision_transformer}

\SetAlgoNlRelativeSize{0}
\KwData{%
Returns-to-go $R$, state $s$, action $a$, timestep $\bar{t}$\;
Linear embedding layers $embed_s, embed_a, embed_R$\;
Learned episode positional embedding $embed_{\bar{t}}$\;
Linear action prediction layer $pred_a$
}

\KwResult{%
Predicted actions for continuous actions
}

\SetAlgoNlRelativeSize{-1}
\textbf{Training phase}\\
\While{not converged}{
    \For{each tuple $(R, s, a, \bar{t})$}{
        Get positional embedding $pos_{emb}$ of timestep $\bar{t}$ using $embed_{\bar{t}}$\;
        Get embed state $s_{emb}$ of state $s$ using $embed_s$\;
        Get embed action $a_{emb}$ of action $a$ using $embed_a$\;
        Get embed returns-to-go $R_{emb}$ of returns-to-go $R$ using $embed_R$\;
        
        Add $pos_{emb}$ to $s_{emb}$, $a_{emb}$, and $R_{emb}$\;

        Stack $R_{emb}$, $s_{emb}$, and $a_{emb}$ sequentially to obtain embed input\;

        Feed embed input to transformer model to obtain hidden states\;

        Extract action tokens from hidden states\;

        Apply action prediction layer $pred_a$ to extracted action tokens to obtain $a_{preds}$\;

        Compute the mean squared error between $a_{preds}$ and actual actions $a$ to obtain model loss\;

        Perform backpropagation and update model parameters;
    }
}

\textbf{Evaluation phase} \\
Set $target~return$ to $1$\;
Initialize $R, s, a, \bar{t}$, done with initial values\;

\While{not done}{
        Get positional embedding ${pos'}_{emb}$ of latest timestep $\bar{t'}$ using $embed_{\bar{t'}}$\;
        Get embed state ${s'}_{emb}$ of latest state ${s'}$ using $embed_s$\;
        Get embed action ${a'}_{emb}$ of latest action ${a'}$ using $embed_a$\;
        Get embed returns-to-go ${R'}_{emb}$ of latest returns-to-go ${R'}$ using $embed_R$\;
        
        Add ${pos'}_{emb}$ to ${s'}_{emb}$, ${a'}_{emb}$, and ${R'}_{emb}$\;

        Stack ${R'}_{emb}$, ${s'}_{emb}$, and ${a'}_{emb}$ sequentially to obtain embed input\;

        Feed embed input to transformer model to obtain hidden states\;

        Extract action tokens from hidden states\;

        Apply action prediction layer $pred_a$ to extracted action tokens to obtain $a_{preds}$\;

        Update $new_s$, $r$, $done$, $\_$ by taking action $a_{preds}$ in the environment\;

        Append the difference between the latest returns-to-go ${R'}$ and $r$ to $R$\;
        Append ${new_s}$ to $s$, append $a_{preds}$ to $a$, append the length of $R$ to $\bar{t}$\;

        Keep the last $K$ elements of $R$, $s$, $a$, $\bar{t}$ for the next iteration.
}
\end{algorithm}

\subsection{Transformer Architecture for GPT-based MADDA}
Transformers, as proposed by \cite{vaswani2017attention}, represent a breakthrough architecture optimized for the efficient modeling of sequential data. These models are composed of a series of stacked self-attention layers interlinked via residual connections. In this algorithm, each self-attention layer processes a set of $w$ embeddings $\{x_o\}_{o=1}^w$, pertaining to distinct input tokens, and produces an output of $w$ embeddings $\{z_o\}_{o=1}^w$, while meticulously preserving the input dimensions.

The core mechanism behind self-attention entails the transformation of each token, denoted by $o$, into three critical components: a key vector $k_o$, a query vector $q_o$, and a value vector $v_o$. These transformations are realized through linear operations. The output signified as $z_o$ for the $o$-th token, is computed by adeptly weighing the values $v_j$ using the normalized dot product of the query $q_o$ with the other keys $k_j$: $z_o = \sum_{j=1}^w softmax({<q_o, k_{j'}>}_{j'=1}^w)_j * v_j$, which enables the self-attention layer to allocate ``credit" by implicitly forming associations between the state and the return, thus maximizing the dot product.

In our study, we employ the GPT architecture \cite{radford2018improving}, a refinement of the transformer architecture specifically tailored for autoregressive generation tasks. Crucially, GPT utilizes a causal self-attention mask, enabling the model to focus solely on previous tokens within the sequence when performing summation and softmax computations ($j \in [1,o]$).

\subsection{GPT-based Deep Reinforcement Learning}
GPT-based DRL is a novel approach for modeling trajectories within the context of offline DRL~\cite{chen2021decision}.
Our method adheres to the core transformer architecture with minimal modifications, as illustrated in Fig.~\ref{fig: GPT_based DRL} and formalized in Algorithm \ref{alg:decision_transformer}. The primary objectives in designing the trajectory representation for GPT-based DRL are two-fold: firstly, to enable the acquisition of meaningful patterns by the transformer model, and secondly to facilitate the conditional generation of actions during test time.

Addressing the challenge of effectively modeling rewards, unlike traditional methods that rely on past rewards for action generation, we aim for our model to generate actions based on the anticipation of future returns. To achieve this, we consider directly feeding the model with rewards and opt instead to provide it with the returns-to-go, denoted as $\hat{R}_t = \sum_{t' = t}^{T} r_{t'}$. This choice results in the following trajectory representation conducive to autoregressive training and generation:
\begin{equation}
\tau = (\hat{R}_1, s_1, a_1, \hat{R}_2, s_2, a_2, \cdots, \hat{R}_T, s_T, a_T) .\label{eq18}
\end{equation}

To encode the trajectory information for processing by the GPT-based DRL, we focus on the most recent $K$ timesteps, with a total of $3K$ tokens, each dedicated to one of the three modalities: return-to-go, state, and action. As shown in Figure \ref{fig: GPT_based DRL}, to derive token embeddings, we utilize linear layers for each modality, transforming raw inputs to the embedding dimension~\cite{chen2021decision}. Layer normalization is then applied. Furthermore, we introduce a unique embedding for each timestep, diverging from the conventional positional embeddings used by transformers. Remarkably, each timestep corresponds to three tokens. The token embeddings are subsequently processed by a GPT model, which employs autoregressive modeling to predict future action tokens.

Algorithm \ref{alg:decision_transformer} presents the pseudo-code for the GPT-based DRL algorithm. In the training phase, we first provide a dataset of offline trajectories. During the training loop, we sample minibatches with a sequence length of $K$. The prediction head corresponding to the input token $s_t$ is tasked with predicting actions $a_t$, utilizing cross-entropy loss for discrete actions and mean-squared error for continuous actions. The accumulated losses for each timestep are subsequently averaged.

In the testing phase, we are afforded the flexibility to specify the desired performance outcome, such as success (encoded as 1) or failure (encoded as 0), along with the initial state of the environment. This information serves as the basis for initiating the generation process. Following this, we implement the generated actions in the current state, reducing the target return based on the received reward until the episode concludes.

For the sequence length of $K$, the self-attention layer has a computation complexity of $\textit{O}(F  K^2  H)$, while the intermediate layer has a computation complexity of $\textit{O}(F  K  H^2)$~\cite{li2022stability}, where $F$ represents the batch size, and $H$ denotes the size of hidden layers. Consequently, the overall computation complexity of GPT-based DRL can be calculated as $\textit{O}(F  K  H  (K + H))$.




\begin{figure}
    \vspace{-0.65cm}
    \centering
    \includegraphics[width=1\linewidth]{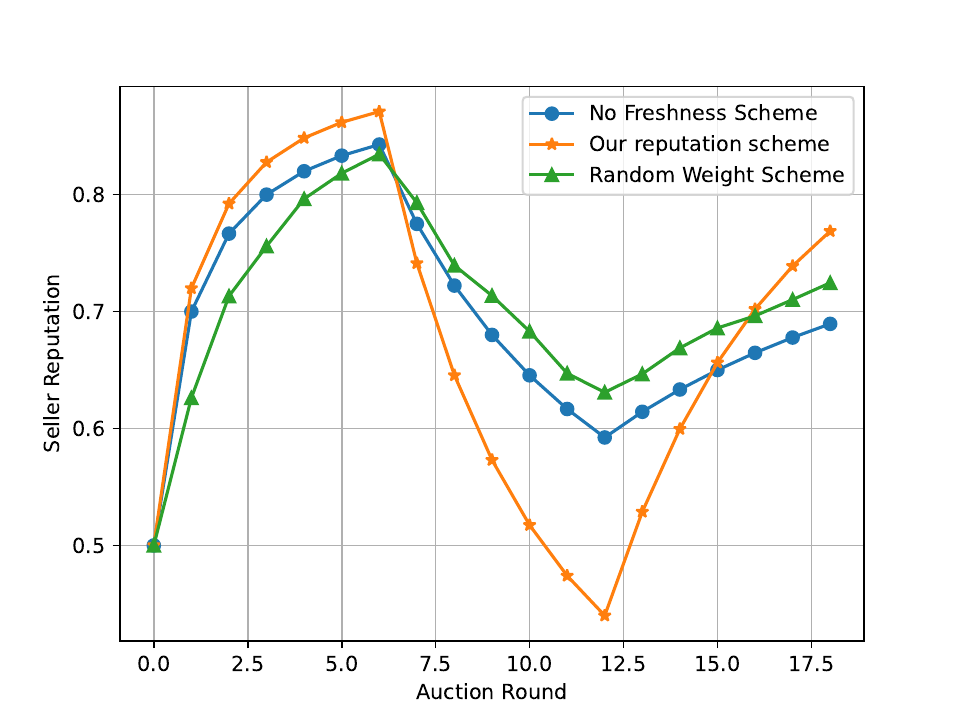}
    \caption{Seller reputation under considering freshness compared with random weight scheme and no freshness scheme}
    \label{fig:enter-label}
\end{figure}

\section{Performance Evaluation}\label{VI}
In this section, we first provide detailed market settings of the vehicular Metaverses. Then, we analyze the convergence performance of the GPT-based auctions. Finally, we conduct more experiments under different settings to validate the performance of the proposed GPT-based auctions.
\subsection{Market Simulation}
\subsubsection{Market Setting}We consider the MADDA market in the vehicular Metaverses, consisting of 50, 100, 150, 200 VUs and 50, 100, 150, 200 RSUs. The required resources $R_{m}$ of VUs to support the VT migration task range from 40 to 80. The maximum distances $q_{m1}$ that VUs expect to migrate from the current RSU to the next RSU range from 0.8 to 1.0 kilometers. The minimum reputation value $q_{m2}$ expected by VUs from the new RSU range from 0.6 to 0.8. The initial reputation value for all RSUs is set to 0.5, which will be updated by the auctioneer with the auction round according to our reputation system. The resources $O_{n}$ owned by RSUs, i.e., computation resources (the number of CPUs) $O_{n}^{cp}$, communication resources $O_{n}^{com}$ (the number of bandwidth resources) and storage resources $O_{n}^{s}$ range from 40 to 80. The parameters $\delta_n$, $\varepsilon$ and $E_{n}$ are set to 0.001, 0.6, and 0.1 respectively. The value of latency sensitivity factor $\hat\alpha$ is set to 0.3. The maximum tolerable latency $\hat{T}$ is set to 0.15 s. In wireless communication, the average transmitter power $\rho$ of RSUs is set to 500 $W$ and the average noise power $N_{0}$ is set to $10^{-9}$ $W/HZ$. The path-loss exponent $\epsilon $ is set to 3. The highest price $p^{max}$ is set to 100, the lowest price $p^{min}$ is set to 1 and the weight $\alpha$ is set to 0.5. The penalty associated with exchange information $\zeta$ is set to 0.01.
\begin{figure}[t]
\vspace{-0.1cm}
\centerline{\includegraphics[width=1\linewidth]{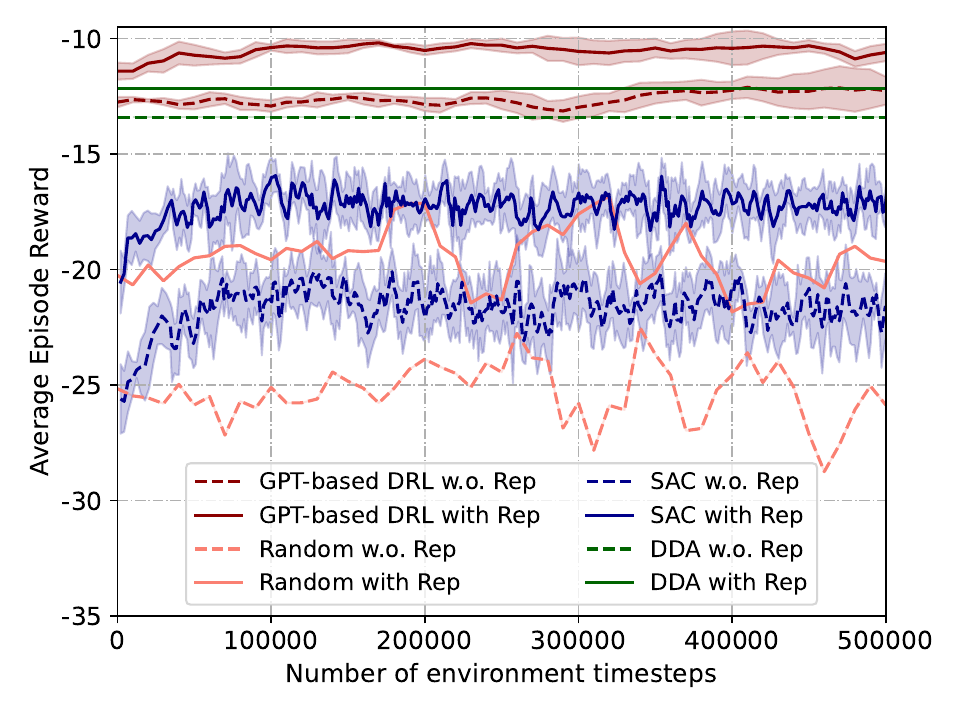}}
\caption{Convergence of the GPT-based DRL compared with DDA, SAC and random methods with or without considering reputation attributes.}
\label{fig4}
\end{figure}

\subsubsection{Comparison Baseline}
\begin{itemize}
\item{Soft Actor-Critic (SAC) method \cite{haarnoja2018soft}:} 
The auctioneer utilizes the exploration and feedback mechanisms of DRL to continuously learn and evaluate the auction environment, dynamically adjusting the Dutch clock and optimizing the clock adjustment strategy, to find a steady and precise convergence of prices.
\item{Random method:} The auctioneer selects the increment for updating the Dutch clock from a range of values, which are distributed evenly.
\item{Traditional Double Dutch Auction (DDA) method:} The auctioneer adjusts the Dutch clock with a fixed single-unit step size. This single-unit step size is the minimum value of the action space in the GPT-based DRL scheme. 
\end{itemize}

\begin{figure*}
  \centering
  \begin{minipage}{0.24\textwidth}
    \subfloat[]{%
      \includegraphics[width=\linewidth]{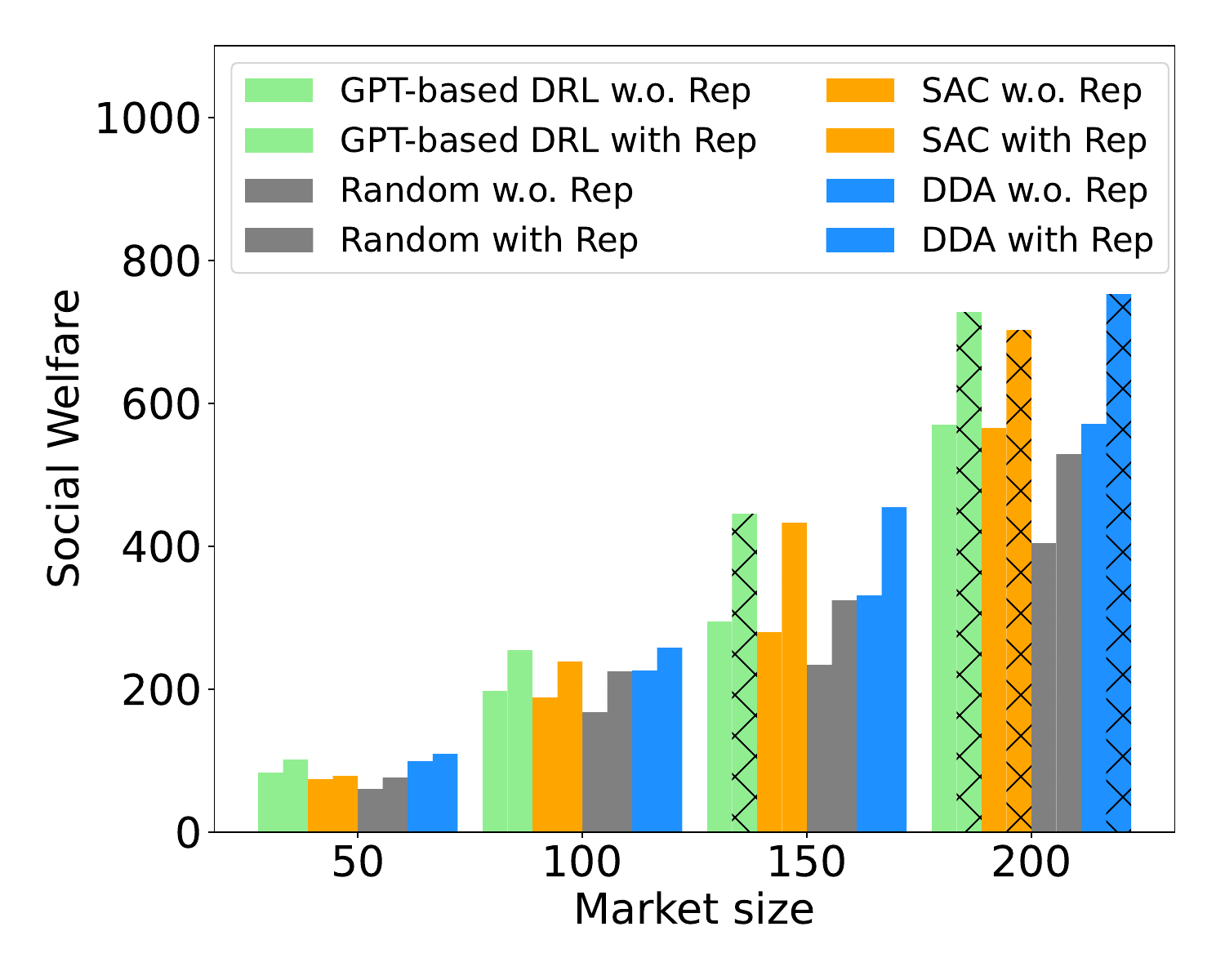}%
    }
  \end{minipage}
  \begin{minipage}{0.24\textwidth}
    \subfloat[]{%
      \includegraphics[width=\linewidth]{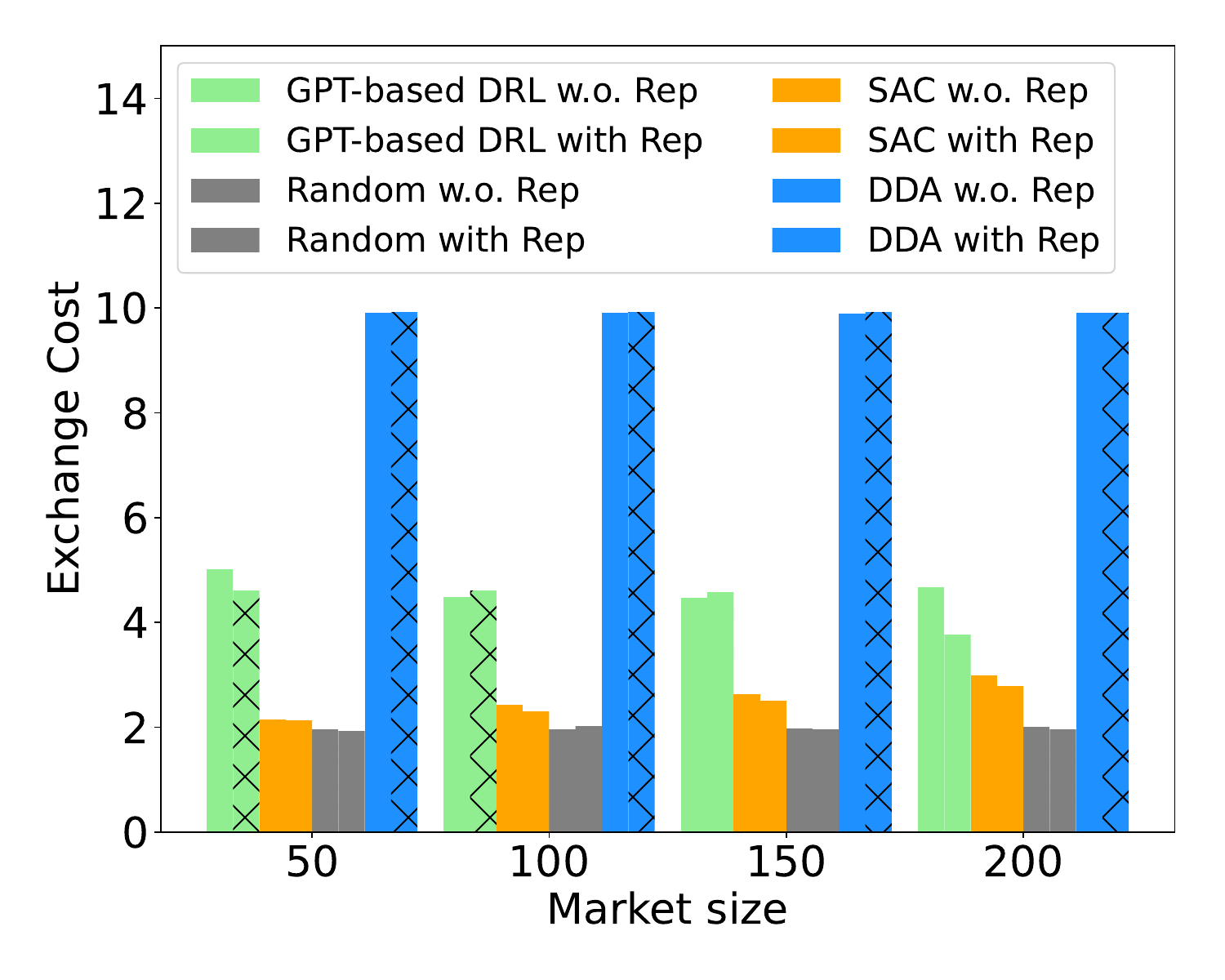}%
    }
      \end{minipage}
  \begin{minipage}{0.24\textwidth}
    \subfloat[]{%
      \includegraphics[width=\linewidth]{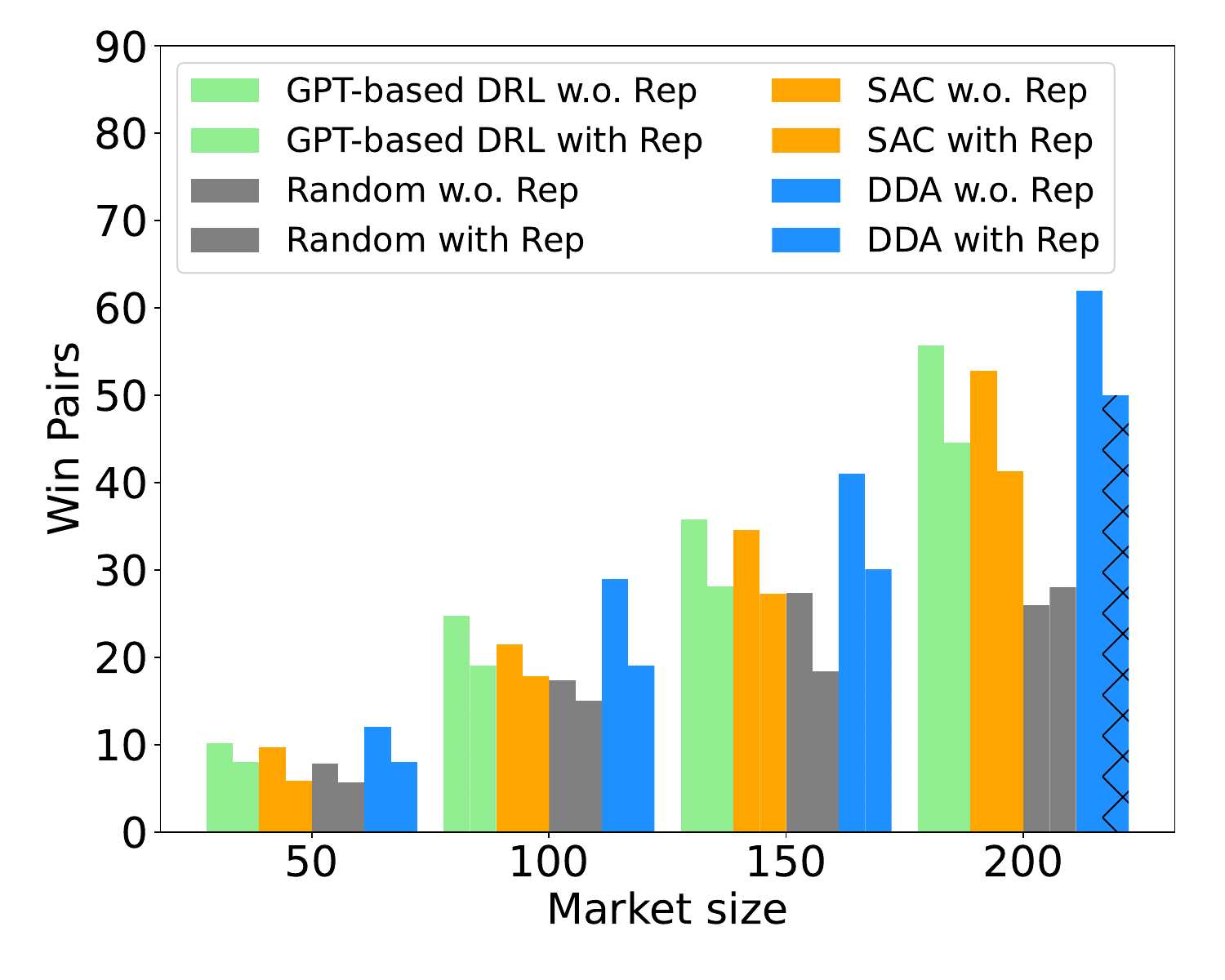}%
    }
      \end{minipage}
  \begin{minipage}{0.24\textwidth}
    \subfloat[]{%
      \includegraphics[width=\linewidth]{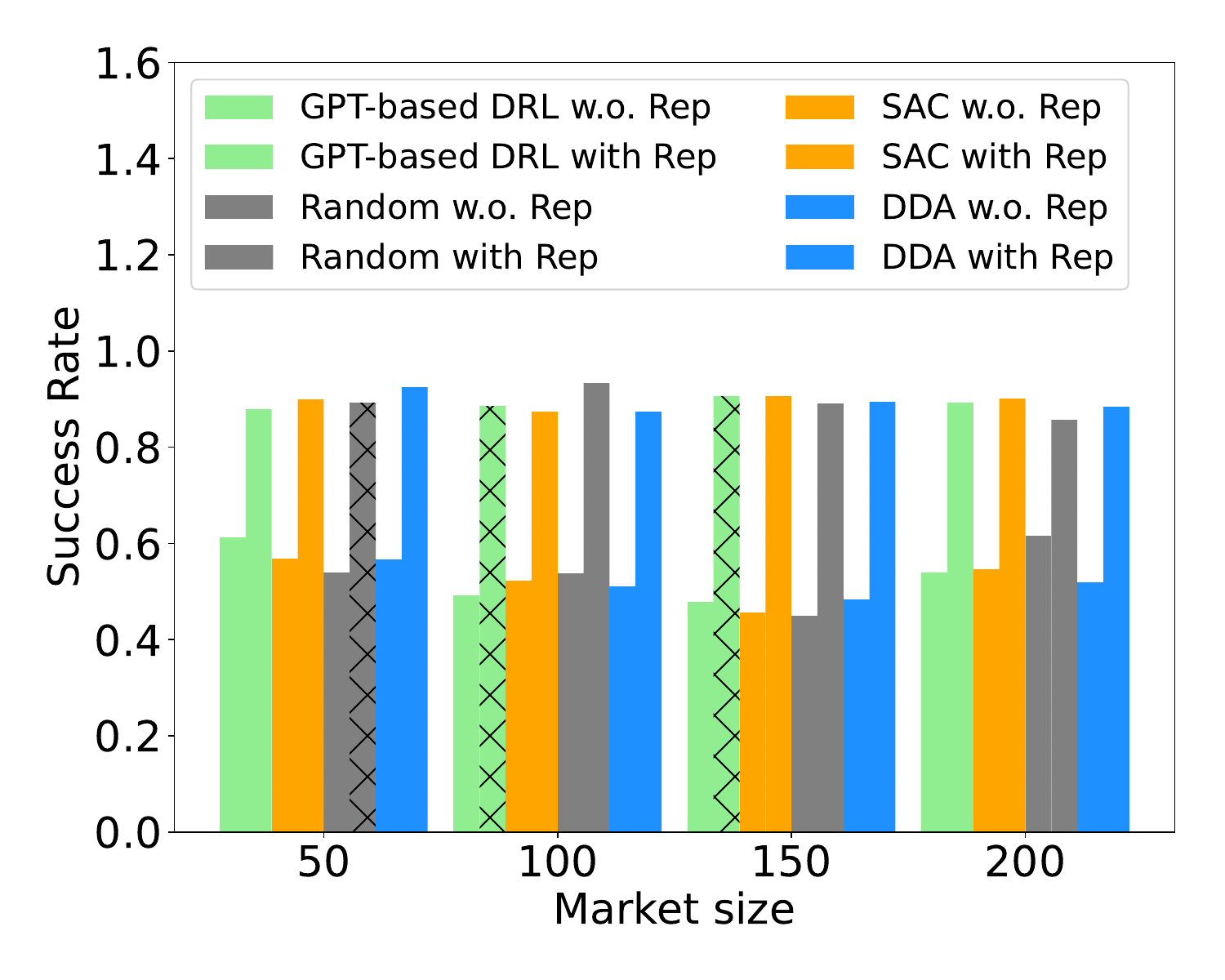}%
    }
  \end{minipage}

  \begin{minipage}{0.24\textwidth}
    \subfloat[]{%
      \includegraphics[width=\linewidth]{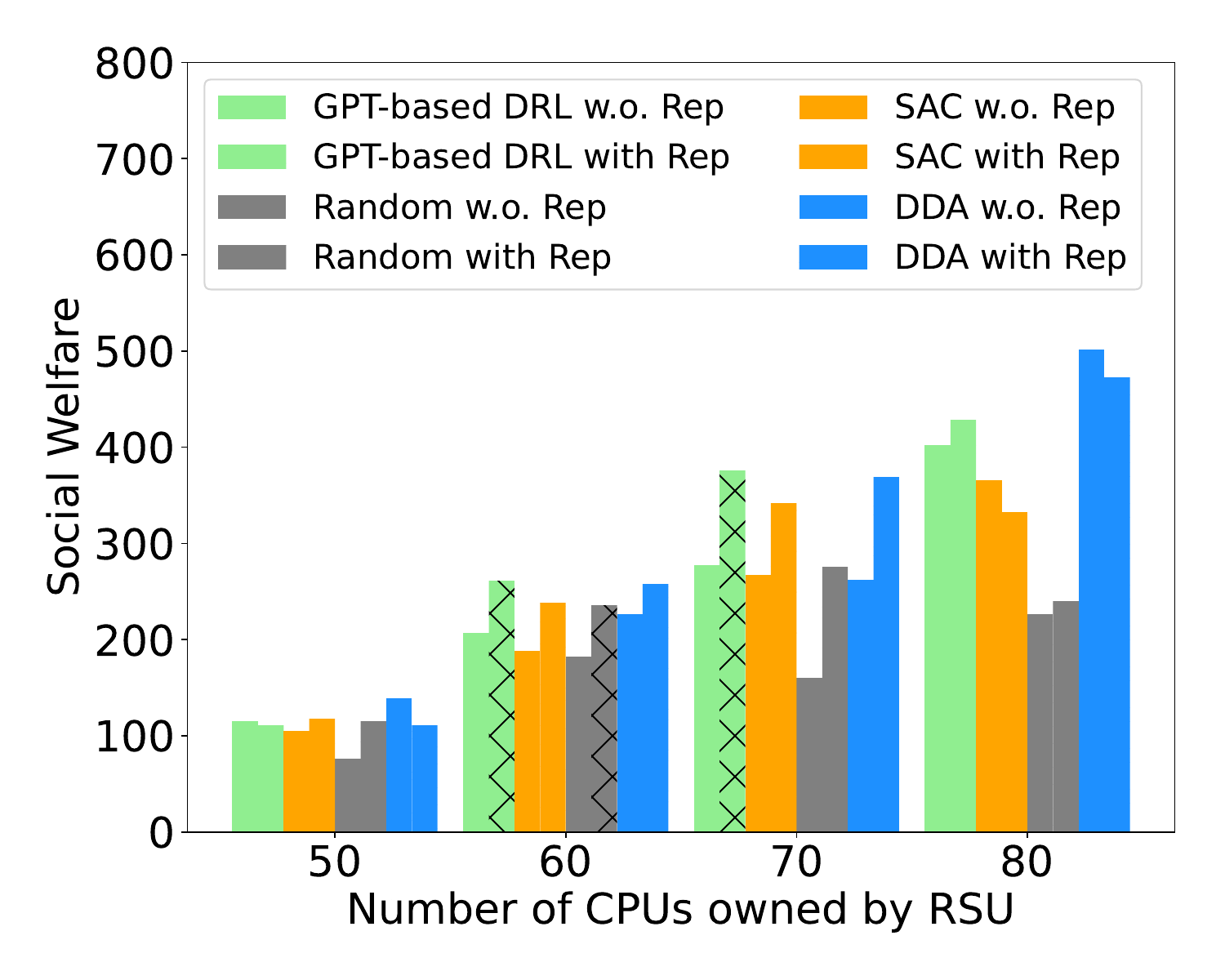}%
    }
    
  \end{minipage}
  \begin{minipage}{0.24\textwidth}
    \subfloat[]{%
      \includegraphics[width=\linewidth]{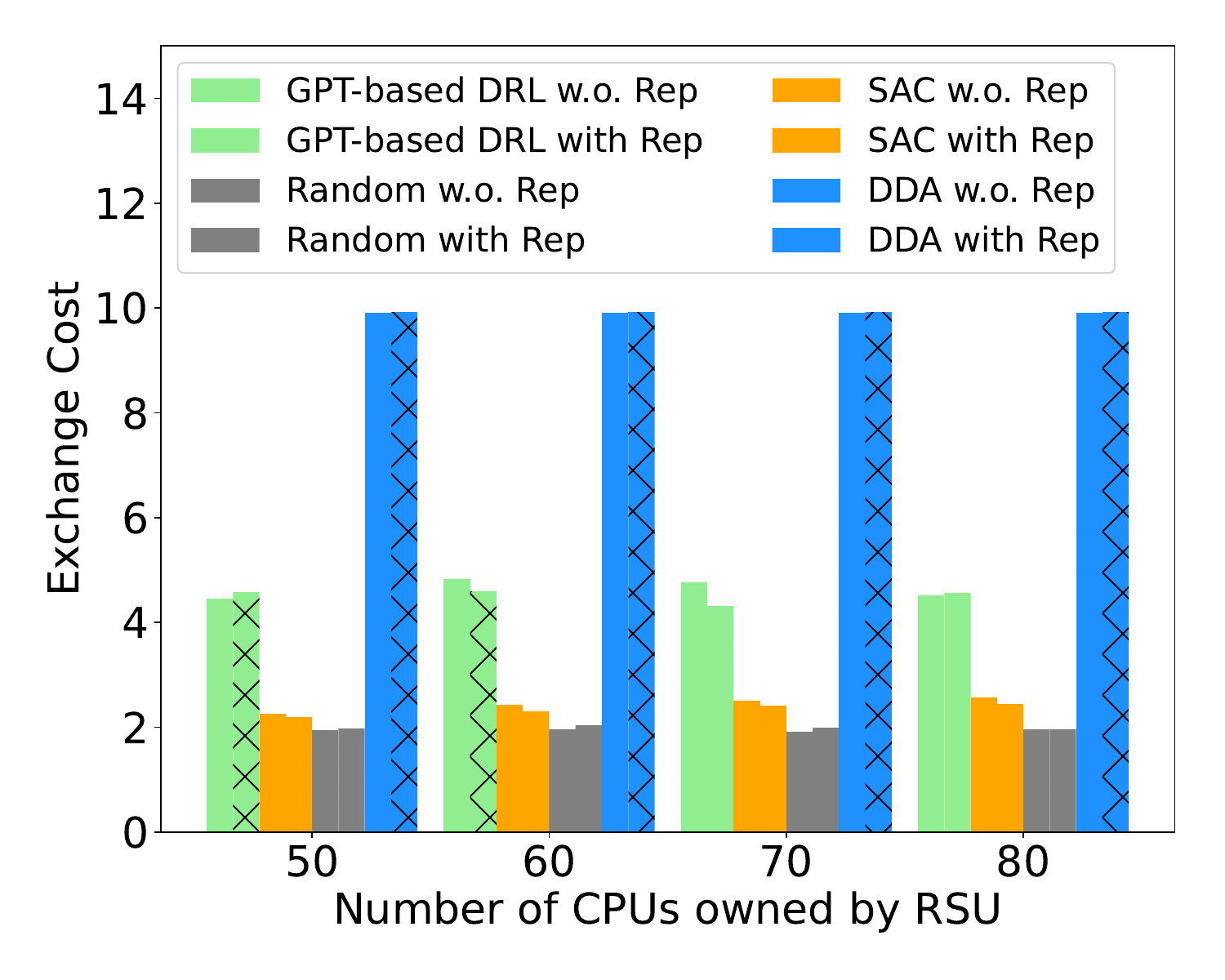}%
    }
  \end{minipage}
  \begin{minipage}{0.24\textwidth}
    \subfloat[]{%
      \includegraphics[width=\linewidth]{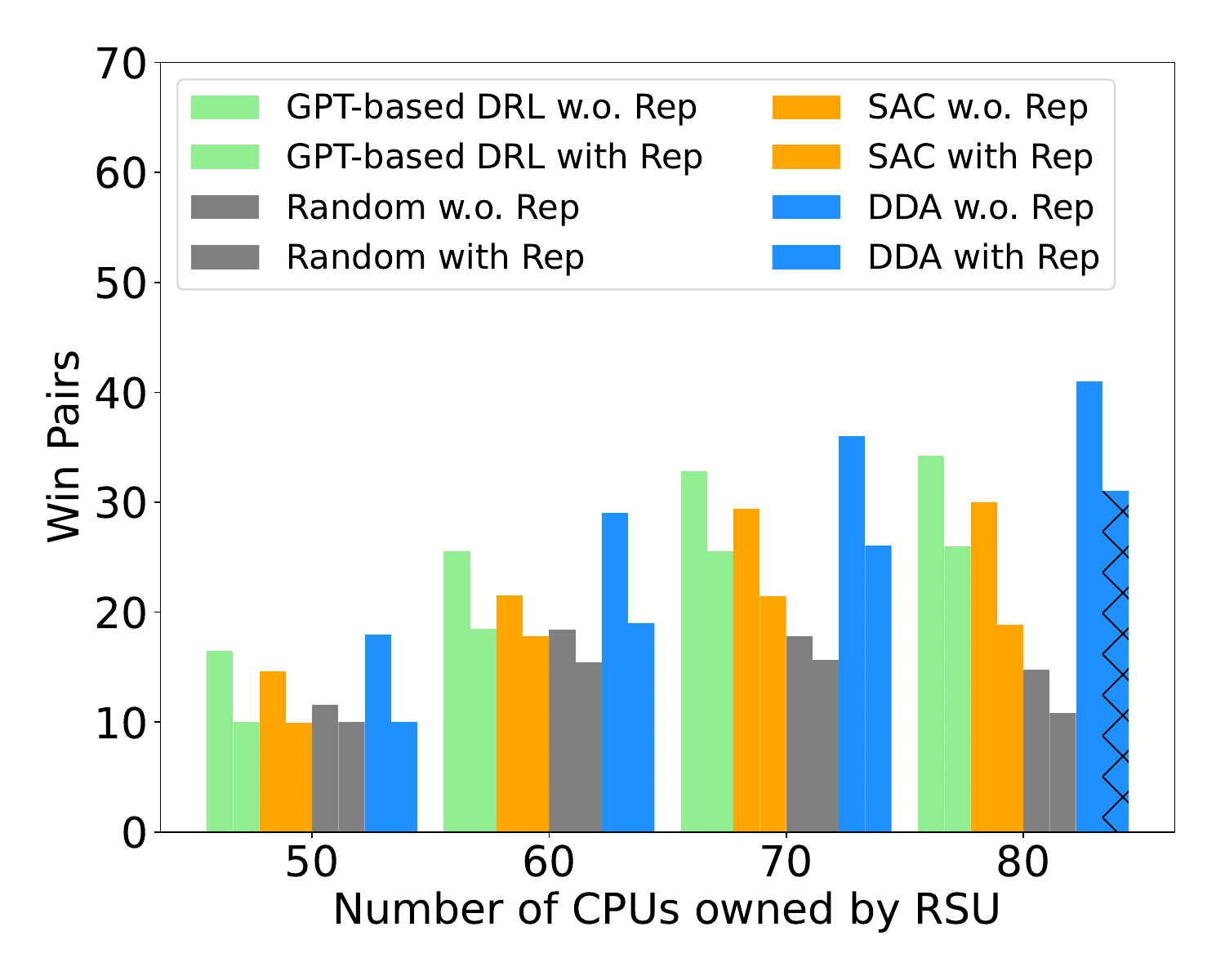}%
    }
  \end{minipage}
    \begin{minipage}{0.24\textwidth}
    \subfloat[]{%
      \includegraphics[width=\linewidth]{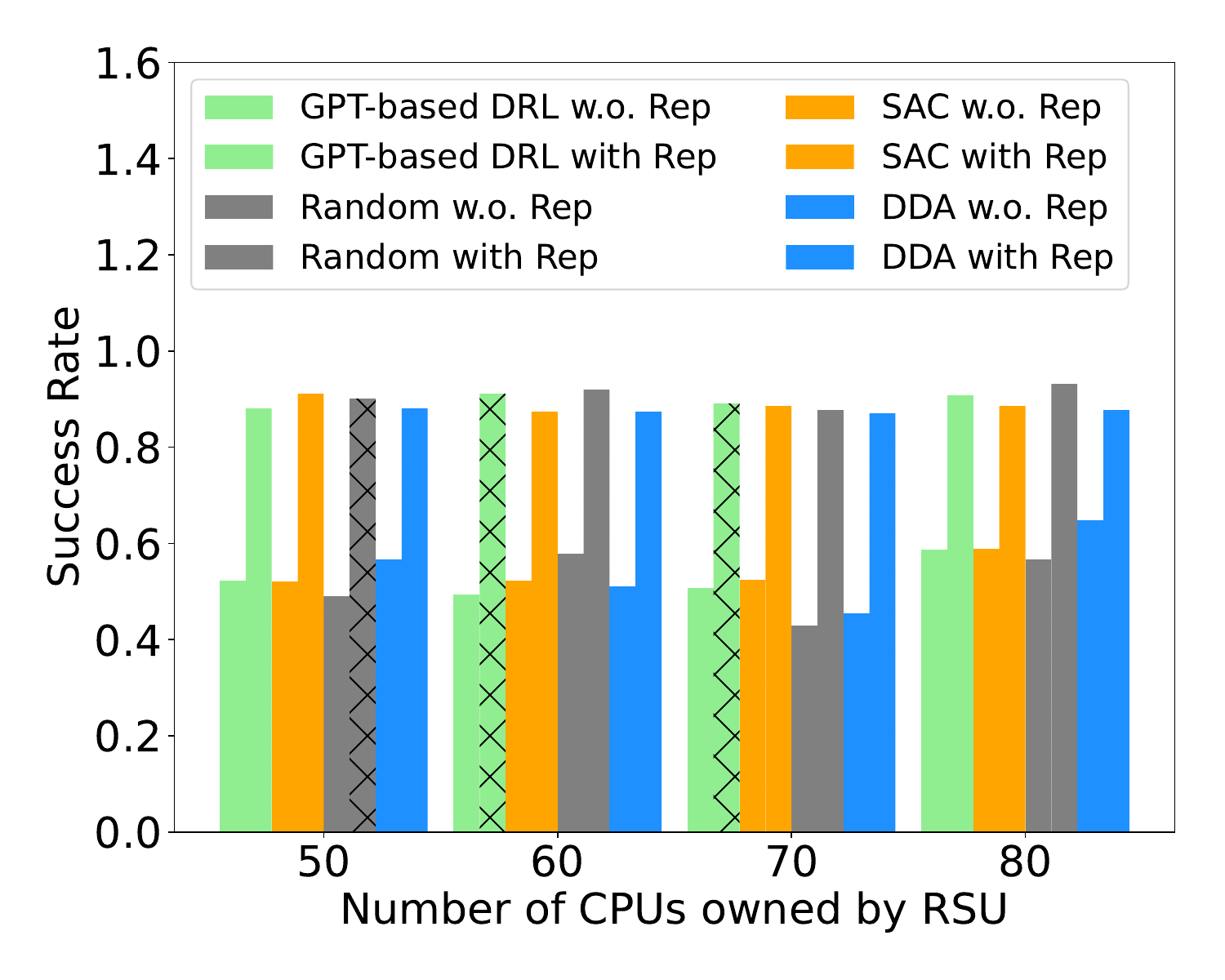}%
    }
  \end{minipage}

  \caption{Performance evaluation under different system parameters. (a) and (b) shows the social welfare and auction information exchange cost versus market sizes. (c) and (d) show the winning pairs and trade success rates under different market sizes, respectively. (e)-(h) show the performance of social welfare, auction information exchange cost, winning pairs, and trade success rates under different computational powers owned by RSUs.}\label{fig5}
\end{figure*}

\subsection{Numerical Results}
This section presents a comprehensive analysis of the results obtained from our experiments. We focus on three key areas: the effectiveness of our freshness-considering reputation system, which considers the freshness of resource feedback evaluations in our reputation model, the convergence efficiency of the MADDA mechanism, and the performance evaluation under various system settings.
\subsubsection{Effectiveness Analysis of the Reputation System} Figure \ref{fig:enter-label} illustrates the response of the reputation system to seller behavior. Initially, a seller's reputation is set to 0.5. Our findings show that if the seller remains honest, their reputation incrementally increases. This upward trend is a testament to the system's effectiveness in acknowledging and valuing integrity, with a steeper rise compared with not considering the freshness scheme and random weighted scheme. Conversely, when a seller shifts from honesty to maliciousness, the reputation system reacts swiftly. Our freshness-focused approach causes a more rapid decline in reputation scores than conventional schemes, underscoring the system's sensitivity and adaptability to changes in seller behavior. This dynamic response of the reputation system is attributed to the system's emphasis on recent user evaluations, allowing for prompt adjustments to the seller's reputation.
\begin{figure}
  \centering
  \begin{minipage}{0.24\textwidth}
    \subfloat[]{%
    \includegraphics[width=\linewidth]{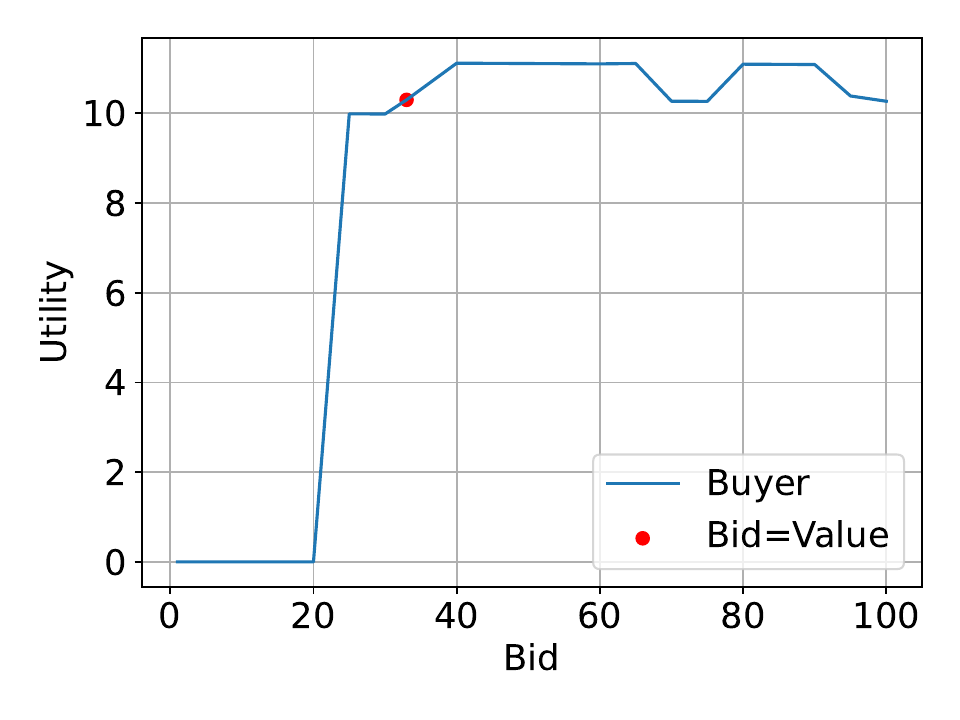}%
    }
  \end{minipage}
  \begin{minipage}{0.24\textwidth}
    \subfloat[]{%
    \includegraphics[width=\linewidth]{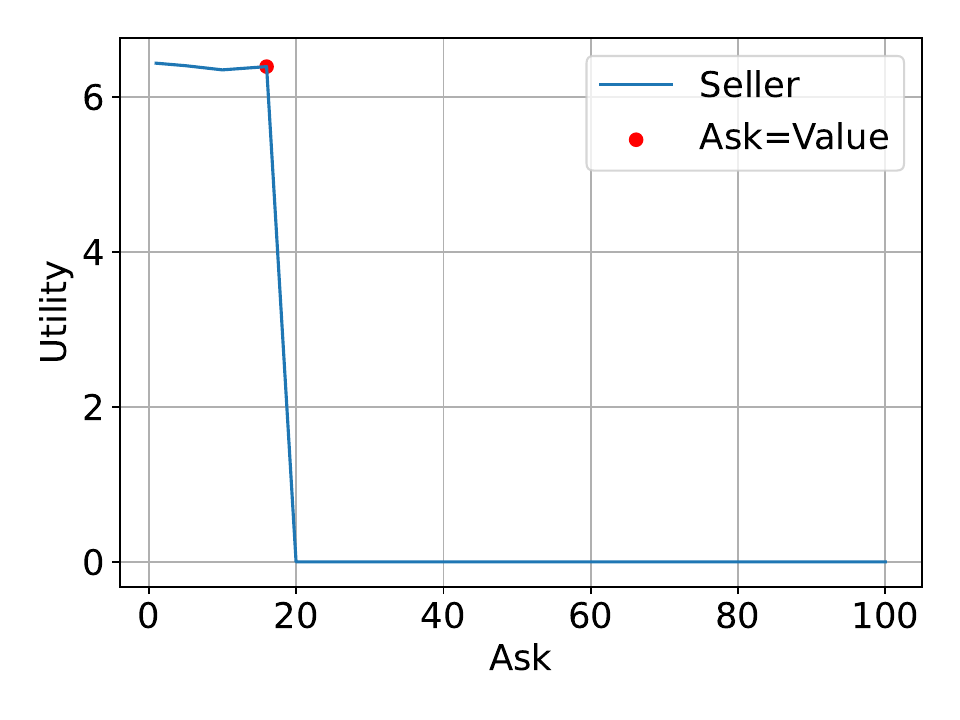}%
    }
      \end{minipage}

  \caption{Individual rationality and incentive compatibility of buyer and seller. (a) for buyer and (b) for seller.}\label{fig IRIC}
\end{figure}

\subsubsection{Convergence Analysis} To demonstrate the efficacy of our MADDA mechanism, which is based on GPT-based DRL, we used several baseline methods for comparison, including the SAC method and the random method. Additionally, to prove that the reputation system we designed can enhance auction efficiency (e.g., social welfare), we also compared the convergence of the auction process when considering reputation versus not considering reputation.

As shown in Fig. \ref{fig4}, during the convergence process, on the one hand, the average reward of the proposed method is higher than that of the other baseline methods. On the other hand, the average reward when considering reputation is also higher than when not considering it. Specifically, Fig. \ref{fig4} demonstrates that, in the case of considering the reputation attribute, the rewards of our scheme are 42\% and 51\% higher than the SAC and the random method, respectively.  Additionally, the auction process converges faster when considering the reputation attribute compared to not considering it. For example, under the GPT-based DRL algorithm, the reward when considering reputation is 18\% higher than when not considering reputation. 

Furthermore, although the DDA method shows stable performance due to using a fixed single-unit step size to adjust the Dutch clock, the fixed step size makes it less adaptable to dynamic auction environments, resulting in higher information exchange costs. In contrast, our proposed GPT-based DRL method can dynamically adjust the clock step size to adapt to changing auction environments, achieving fast convergence and near-optimal social welfare with low information exchange costs.

\subsubsection{Performance Evaluation under Different Setting} To verify the generalization ability of the proposed GPT-based MADDA mechanism, we conduct tests under different system parameter settings, e.g., the sizes of the multi-attribute resource market and the computational capabilities of RSUs (only consider the number of CPUs owned by RSU), with other baseline methods.

In Figs. \ref{fig5}(a) and \ref{fig5}(b), we compare the social welfare and auction information exchange costs achieved with and without considering reputation attributes in different market sizes (i.e., the number of VUs and RSUs) against other baseline methods. As shown in Fig. \ref{fig5}(a) and \ref{fig5}(b), social welfare and exchange costs increase with the size of the market. This is because the larger the market size, the more intense the participation during the auction, while the number of winning pairs increases, as Fig. \ref{fig5}(c) showed. Furthermore, from Fig. \ref{fig5}(a) and \ref{fig5}(b), it can be observed that the GPT-based DRL method, considering reputation attributes, achieved near-optimal social welfare with lower information exchange cost across different market sizes compared to the traditional DDA method. Under the condition of considering reputation attributes, the social welfare achieved by the GPT-based DRL method is average 9.2\% and 22.8\% higher than the SAC method and the random method, respectively, demonstrating that the GPT-based DRL method significantly outperforms the random method in terms of social welfare. Additionally, the performance in social welfare varies depending on whether reputation attributes are considered. Compared to not considering reputation attributes, considering them leads to higher social welfare and lower information exchange costs. This is because considering reputation attributes improves the transaction quality of matched pairs and reduces ineffective pairings, thus enhancing social welfare. 

In addition, we use the number of winning pairs and success trade rate to measure the system performance metrics under different market sizes. From Fig. \ref{fig5}(c), it can be
observed that our proposed GPT-based MADDA mechanism always has more winning pairs than other baseline schemes. We can see that considering the reputation attribute can drastically improve the transaction success rate from Fig. \ref{fig5}(d).

Finally, we compare the performance of different computational powers of the RSUs at the same market size under different baselines, such as social welfare, information exchange costs, winning pairs, and trade success rate. As can be seen from Fig. \ref{fig5}(e)-(h), our proposed GPT-based MADDA mechanism performs better in terms of social welfare, auction information exchange costs, winning pairs, and trade success rates compared to the other baselines under different computation powers of RSUs. Specifically, Figs. \ref{fig5}(e) and \ref{fig5}(f) show the trend of social welfare and auction information exchange cost, as the computational powers of the RSUs increase under different baselines. From Fig. \ref{fig5}(e), the social welfare in the market increases as the computational powers of the RSUs increase. The reason is that as the computation powers of the RSUs increase, making RSUs more competitive in the marketplace and potentially generating more winning pairs, while the quality of the VTs migration tasks also increases, the utility of the RSUs will increase.
\subsubsection{Incentive Compatibility (IC) and Individual Rationality (IR)} To demonstrate the IC and IR~\cite{dobzinski2012multi} properties of the proposed multi-attribute double Dutch auction mechanism, we conduct the following experiment. We fix the market size and randomly select one buyer and one seller from the market. All experimental parameters are kept constant except for the buy-bid and the sell-bid. We vary the buy-bid and the sell-bid, including values both greater than and less than their true values, ranging from $p^{min}$ to $p^{max}$. 
In Figs.~\ref{fig IRIC}(a) and~\ref{fig IRIC}(b), the red dots represent the points where the bid equals the value and the ask equals the value, respectively, highlighting the scenarios of truthful bidding and asking. From 
Fig.~\ref{fig IRIC}(a), the utility of the buyer increases as the bid approaches the true value, reaches a maximum when the bid equals the true value, and then decreases as the bid exceeds the true value. This indicates that the buyer’s optimal strategy is to bid truthfully, which demonstrates IC for buyers. The fluctuations in the buyer’s utility are due to the DRL executing non-fixed actions, which are influenced by the bid, causing the price to vary and resulting in these fluctuations. 
From Fig.~\ref{fig IRIC}(b), the utility of the seller decreases as the ask diverges from the true value. The utility is maximized when the ask equals the true value. This shows that the seller’s optimal strategy is to ask truthfully, which demonstrates IC for sellers. From Figs.~\ref{fig IRIC}(a) and~\ref{fig IRIC}(b), it can be observed that as the buy-bid and sell-bid increase, the utilities of both the buyer and the seller always remain greater than or equal to zero, thereby validating IR.


%


\section{Conclusions}\label{VII}
In this paper, we introduced a novel GPT-based incentive mechanism in the multi-attribute resource market for VTs migration to efficiently allocate multi-dimensional resources between VUs and RSUs in vehicular Metaverses. Our approach uniquely considers both price and non-monetary attributes, employing a two-stage matching process that first achieves optimal resource-attributes matching, then followed by a winner determination and pricing strategy using a GPT-based DRL auctioneer to adjust the auction clocks efficiently during the DDA process. Our experimental results demonstrated the effectiveness of this mechanism in enhancing convergence efficiency and achieving near-optimal social welfare with lower information exchange cost. This approach also adapts well to dynamic seller behaviors through a freshness-considering reputation system, while performing better in terms of winning pairs and trade success rates with similar exchange cost compared to the other baselines.
To implement this scheme in real-world applications, vehicular networks can integrate our GPT-based auction mechanism into their resource management systems. RSUs with edge computing capabilities can dynamically allocate resources to vehicles needing VT migration by learning and adapting through real-time data collection and processing.
Future work will enhance the GPT-based DRL algorithm's ability to predict and respond to environmental changes, ensuring efficient resource allocation and seamless experience of users.

\ifCLASSOPTIONcaptionsoff
  \newpage
\fi



%

 \bibliographystyle{IEEEtran}
\bibliography{main}

%






\end{document}